\pdfoutput=1

\documentclass[11pt]{article}

\usepackage[final]{ACL2023}

\usepackage{times}
\usepackage{latexsym}

\usepackage[T1]{fontenc}

\usepackage[utf8]{inputenc}

\usepackage{microtype}

\usepackage{inconsolata}
\usepackage{tabularray}
\usepackage{graphicx}
\usepackage{multirow}
\usepackage{booktabs}
\usepackage{adjustbox}
\usepackage{enumerate,enumitem}
\usepackage{xcolor, soul}
\usepackage{ulem}
\usepackage{fontawesome5}

\sethlcolor{lightgray}

\useunder{\uline}{\ul}{}

\title{Investigating the Impact of Model Instability on Explanations and Uncertainty} %

\author{Sara Vera Marjanović\qquad Isabelle Augenstein \qquad Christina Lioma \\   Department of Computer Science \\ %
University of Copenhagen\\ 
\texttt{\{savema, augenstein, c.lioma\}@di.ku.dk}}

\begin{document}

\maketitle

\begin{abstract}
Explainable AI methods facilitate the understanding of model behaviour, yet%
, small, imperceptible perturbations to inputs can vastly distort explanations. As these explanations are typically evaluated holistically, before model deployment, it is difficult to assess when a particular explanation is trustworthy. %
Some studies have tried to create confidence estimators for explanations, but none have investigated an existing link between uncertainty and explanation quality. We artificially simulate epistemic uncertainty in text input by introducing noise at inference time. In this large-scale empirical study, we insert different levels of noise perturbations and measure the effect on the output of pre-trained language models and different uncertainty metrics.
Realistic perturbations have minimal effect on performance and explanations, yet masking has a drastic effect. We find that high uncertainty doesn't necessarily imply low explanation plausibility; the correlation between the two metrics can be moderately positive when noise is exposed during the training process. This suggests that noise-augmented models may be better at identifying salient tokens when uncertain. \color{black} Furthermore, when predictive and epistemic uncertainty measures are over-confident, the robustness of a saliency map to perturbation can indicate model stability issues. Integrated Gradients shows the overall greatest robustness to perturbation, while still showing model-specific patterns in performance; however, this phenomenon is limited to smaller Transformer-based language models.
\end{abstract}

\vspace{0.5em}
    \hspace{.5em}\faGithub\hspace{.75em}\parbox{\dimexpr\linewidth-2\fboxsep-2\fboxrule}{\url{https://github.com/spaidataiga/unc-and-xai-noise}}
    \vspace{-.5em}

\color{black}
\section{Introduction}

Though language models have become increasingly popular for personal and industrial use, %
these black-box models have been prone to perpetuate discrimination and output hallucinations \cite{augenstein2023factuality,bang2023multichatgpt,Weidinger2021EthicalModels}. To use these models safely, it is important to instil a level of trust in their output. Some methods of instilling trust in a model output include 
\textit{uncertainty} estimation and \textit{eXplainable AI} (XAI). Uncertainty is a reflection of a model's confidence in its output, given, for example, ambiguous or unfamiliar data. While uncertainty can be estimated at inference time in an unsupervised manner, XAI is typically holistically evaluated for a model and task  \cite{Chen2022WHATEXPLANATIONS,Hedstrom2023Quantus:Beyond}. However, XAI techniques give unstable explanations given small changes in input data \cite{Adebayo2018SanityMaps,Alvarez-Melis2018OnMethods,Lakkaraju2020HowExplanations}. While these studies have been critiqued for inserting unnatural noise into the input data, even relatively realistic perturbations to images can disrupt most gradient-based saliency map techniques \cite{Amorim2023EvaluatingPerturbations}. 

\begin{table*}[]
\centering
\small
\begin{tabular}{@{}ll@{}}
\toprule
\multicolumn{1}{c}{\textbf{Noise type}} & \multicolumn{1}{c}{\textbf{Example text}}                                                  \\ \midrule
\multicolumn{1}{c}{(unperturbed)}      & “an artful intelligent film that stays within the confines of a well-established genre”    \\ \midrule
\multicolumn{1}{l}{token-MASK}               & “an \hl{{[}MASK{]}} \hl{{[}MASK{]}} film that stays within the confines of a \hl{{[}MASK{]}} genre”       \\
\multicolumn{1}{l}{token-UNK}                & “an \hl{{[}UNK{]}} \hl{{[}UNK{]}} film that stays within the confines of a \hl{{[}UNK{]}} genre”          \\
\multicolumn{1}{l}{charinsert}         & “an \hl{artfuVl} \hl{intDelligent} film that stays within the confines of a \hl{well-Mestablished} genre” \\
\multicolumn{1}{l}{charswap}           & “an \hl{artfjl} \hl{intellhgent} film that stays within the confines of a \hl{Pell-established} genre”    \\
\multicolumn{1}{l}{butterfingers}      & “an \hl{artdul} \hl{intelligegt} film that stays within the confines of a \hl{well-esfablished} genre”    \\
\multicolumn{1}{l}{l33t}               & “an \hl{@r7fu1} \hl{1n7311193n7} film that stays within the confines of a \hl{w311-357@611543d} genre”    \\
\multicolumn{1}{l}{synonym}            & "an \hl{disingenous} \hl{sound} film that stays within the confines of a \hl{good-established} genre"     \\ \bottomrule
\end{tabular}
\caption{All 7 types of perturbation visualized on a datapoint where 25\% of human-salient tokens are perturbed}
\label{tab:noises}
\end{table*}

Due to this instability, it is difficult to know when we can trust a specific explanation%
. Ideally, we would like to use XAI to understand both why a model succeeds and fails to identify points of failure in a model pipeline-- these failures could arise from mistakes in the model training or ambiguity within the data.
It is also vital to understand when explanations are trustworthy, as the inclusion of XAI can cause an over-reliance on models \cite{Bauer2023PleaseKnowledge,vanderWaa2021EvaluatingExplanations}, give users the false impression of global task understanding \cite{Chromik2021IAI}, and lead to overall poorer performance than if no human-AI collaboration \cite{Schmidt2020TransparencySystems}. Therefore, we would like to assess if the uncertainty of a model's output can give any indication of an explanation's quality and if the instability of an explanation can provide insight into the model's performance. %
We expect noise at inference time, especially for text: Words can be accidentally ablated, mispelled or otherwise mutated. Different authors have distinct linguistic styles, and new words emerge or change in meaning. Due to this noise, many SOTA language models suffer out-of-distribution issues and, thus, fail in real-world applications \cite{Alipanahi2022UnderspecificationLearning, Ribeiro2020BeyondCheckList}. As large language models rely on drawing from large amounts of data (often stemming from sources with variable writing styles and formatting, like social media), we must understand how this ``noise'' in the data affects model's performance, confidence, and explainability. As text perturbations can introduce some ambiguity into the data that is not present at training time, they should affect a model's reported uncertainty alongside its explanation. Given the variety of language models available, it is also vital to compare how this relationship differs across different models and XAI methods.

In this paper, we conduct a large-scale empirical investigation into the effect of noise on Pre-trained Language Models via a controlled experiment by artificially injecting varying degrees and types of realistic noise (see Table \ref{tab:noises}) and measuring the impact on model explanations and uncertainty. In this manner, we also investigate the relationship between explanation plausibility (the agreement between model saliency and ground-truth annotations) and model uncertainty. To assess if explanation instability reflects model instability, we limit our investigation to gradient-based techniques, given their high-performance in robustness and plausibility measures \cite{Atanasova2020AClassification} and to limit additional uncertainty introduced by model approximation techniques, like LIME \cite{Zhang2019WhyExplanations}.

Here, we provide the following \textbf{contributions}:

\begin{itemize}[noitemsep] %
    \item We evaluate, for the first time, the relationship between uncertainty and explanation plausibility given perturbed and unperturbed data;
    \item We assess on a large-scale how the degree of artificial noise at inference time affects model performance, confidence and explanation plausibility across a variety of transformer-based language models, degrees of perturbation, and methods of perturbation;
    \item We compare four popular XAI methods in their robustness to noise across noise types and models at different levels of perturbation.
\end{itemize}

We find that high uncertainty does not imply low explanation plausibility; models trained with noisy data can still generate coherent explanations despite high uncertainty amid noise. Furthermore, we argue that explanation instability can give some insight into model performance and can show patterns in saliency attribution: Common, realistic perturbations (like synonym replacement) have smaller effects on model performance and saliency maps, yet l33t speak and token replacement have a larger impact. This pattern is seen typically strongest in Integrated Gradients, which also shows the greatest robustness for smaller language models.

\section{Related Work}

\paragraph{Assessing trustworthiness}
There are many ways to assess a model's trustworthiness for a task or inference. The confidence in an output can be quantified via its uncertainty, and the reasonability of an output can be assessed via XAI. Furthermore, the overall quality of an XAI method can be evaluated, either via the similarity to human annotations or via other metrics like robustness to noise or conciseness \cite{Hedstrom2023Quantus:Beyond,Chen2022WHATEXPLANATIONS, Atanasova2020AClassification}. There is some controversy within these measures: Models that output explanations with high similarity to human-annotations may result in unfaithful explanations, as models may not actually rely on this information to compute their output \cite{Jin2023RethinkingPlausibility}. %
Moreover, these explanations can also be unstable and prone to large changes in output given small changes in input data \cite{Adebayo2018SanityMaps,Alvarez-Melis2018OnMethods,Lakkaraju2020HowExplanations,Hedstrom2023Quantus:Beyond,Chen2022WHATEXPLANATIONS}. However, these (often image) studies do not investigate the causes of the instabilities or how they relate to other measures, like uncertainty. %

\paragraph{Noise on language model performance}
Several other studies have looked specifically at the effect of noise on the performance and confidence of BERT-related models. Surprisingly, there are %
contrasting effects of noise on machine and human ability to perform natural language understanding tasks. Perturbations that do not affect a human's ability to understand text significantly perturb BERT performance \cite{Jin2019IsEntailment, Wang2022SemAttack:Spaces}, yet perturbations that worsen human performance do not affect model performance \cite{Feng2018PathologiesDifficult, Gupta2021BERTUnderstanding,Sinha2021UnNaturalInference}. The impact of different kinds of noise differs across model types \cite{moradi-samwald-2021-evaluating}, and the more ``learnable'' a kind of noise is for a model, the less performance decays given noise-augmented data \cite{Zhang2022InterpretingPerturbations}. However, as these studies focus on BERT-related models, there is limited focus on other model types, like GPT, and they also do not evaluate explanations.

\paragraph{Uncertainty measures} 
The `learnability' of a trait or type of noise can be likened to \textit{epistemic uncertainty}, which is a measure of uncertainty in a model's parameters. This is believed to be malleable given more training time and data \cite{Gal2015DropoutLearning}.  %
In contrast, \textit{aleatoric uncertainty} stems from noise inherent in the data generation process \cite{Kendall2016WhatVision}. Many studies conflate the two forms of uncertainty by only looking at the softmax of the output logits as a measure of confidence (hereon named \textit{predictive uncertainty}). However, these measures can be prone to over-confidence%
. For example, when provided highly perturbed data, model confidence increases, even with the addition of calibration methods \cite{Feng2018PathologiesDifficult,Gupta2021BERTUnderstanding}. As these studies use the conflated measure of predictive uncertainty, it is difficult to ascertain the cause of this confidence increase. Therefore, we include epistemic uncertainty as a measure in our study. %

\paragraph{Uncertainty and XAI} 
Other works in the intersection of uncertainty and XAI quantify the uncertainty of a given explanation, by developing new models \citep{Bykov2020HowNetworks} or looking at ensemble explanations \cite{Chai2018UncertaintyInterpretability, Slack2020ReliableExplainability,Marx2023ButAI}, or they attempt to explain the causes of a model's uncertainty \cite{Brown2022UsingUncertainty,Watson2023ExplainingValues}. In \citet{Marx2023ButAI}, they find that the size of the dataset is inversely proportional to the uncertainty of the explanations, which suggests that, with increased training data, XAI techniques tend to converge; therefore, epistemic uncertainty may affect XAI explanations. However, these methods do not look at existing links between XAI and uncertainty and look mainly at image and synthetic datasets.

\begin{table}
\small
\centering
\begin{adjustbox}{max width=1.0\textwidth,center}
\begin{tblr}{
  hline{1,5} = {-}{0.08em},
  hline{2} = {-}{},
}
\textbf{Dataset}       & \textbf{Task}                & \textbf{Size}                           \\
{SemEval 2013\\Task 2} & {Sentiment\\ Classification} & {Training: 4133\\ Annotated Test: 1659} \\
{SST-2 +\\Hummingbird} & {Sentiment\\ Classification} & {Training: 67349\\ Annotated Test: 62}  \\
HateXplain             & {Hatespeech\\Detection}      & {Training:~15383\\Annotated Test:~1142}  \\

\end{tblr}
\end{adjustbox}
\caption{Our training and test datasets. We restrict our test datapoints to those including human-annotated explanations (`Annotated Test').}
\label{tab:Datasets}
\end{table}

In summary, most studies investigating noise on model output look only at small levels of perturbation and focus on a small subset of language models. Furthermore, they conflate different sources of uncertainty in their investigation and do not assess their link to saliency attribution. In our paper, we investigate the effect of different scales of perturbations on a range of popular language models, including GPT2 and OPT. In addition, to avoid conflating sources of uncertainty, we use multiple measures of uncertainty to assess the relationship between model instability and saliency attribution. %

\section{Methods}

\subsection{Datasets}

We limit relevant tasks and datasets for this investigation to publicly available English datasets. We select simple, popular text classification tasks %
with text that has been annotated for importance at word-level granularity by multiple (2+) annotators. We summarize the datasets in Table \ref{tab:Datasets}. Within sentiment classification, we have two datasets: Hummingbird \cite{Hayati2021DoesLexica} and the Semeval-2013 Task 2 dataset \cite{Nakov2013SemEval-2013Twitter}. Hummingbird is a re-annotated subset of several datasets, including the SST-2 dataset \cite{socher-etal-2013-recursive}. We restrict the Hummingbird Sentiment test dataset to only datapoints originating from the SST-2 validation set and train on the SST-2 train dataset. %
We remove neutral datapoints from SemEval-2013 dataset and HateXplain \cite{Mathew2020HateXplain:Detection} to avoid issues of the sufficiency of highlighted text as explanations \cite{Wiegreffe2021TeachProcessing}. %

\subsection{Models}
We test the performance of five different open-source large pre-trained language models: BERT\textsubscript{base} \cite{Devlin2018BERT:Understanding}, RoBERTa\textsubscript{base} \cite{Liu2019RoBERTa:Approach}, ELECTRA \cite{Clark2020ELECTRA:GENERATORS}, GPT-2\textsubscript{medium} \cite{Radford2019LanguageLearners}, and OPT-350M \cite{ZhangOPT:Models}, chosen due to their variety in pretraining and their popularity. We describe their finetuning in Appendix \ref{sec:APP_searchspace}.

\subsection{Perturbations}
\label{sec:perturb}

At test time, we introduce varying levels, hierarchies, and types of perturbations to simulate epistemic uncertainty. A singular type of perturbation is applied to space-delimited words following different hierarchies for increasing \textbf{levels}, or proportions, of the text ($\alpha\in \{0, .05, .10, .25, .50, .70, .80, .90, .95\})$; more details are in Appendix \ref{sec:APP_prop}.

We use three \textbf{hierarchies} of preferential perturbation: \verb|random|, \verb|human|, and \verb|gradient|. Random-hierarchy is determined randomly, though the pattern of perturbed words is preserved across increasing levels of perturbation. Human-hierarchy is determined by the word-level annotations of the dataset. Non-annotated words are then ranked via their part-of-speech tag. We assess the efficacy of this perturbation approach in Appendix \ref{app:RvsS}. %
Gradient-hierarchy is calculated specific to each model as it is ranked by words with the greatest average change according to the Hotflip candidates table \cite{Ebrahimi2018HotFlip:Classification}. When combining tokens to create full words, we take the mean of token gradients. This was determined after taking a subsample of the datapoints and choosing the aggregation method giving the lowest mean ranking to NLTK stopwords. %

We introduce %
seven different noise \textbf{types} to the datapoints (see Table \ref{tab:noises}), selected from previous work in text perturbation: At a fine-grained level, we introduce a random character into a random section of the word (\verb|charinsert|), randomly replace a character in a word (\verb|charswap|) or replace a random character with a character nearby on a qwerty keyboard (\verb|butterfingers|). These insertions have been implemented in other studies on adversarial perturbation in text \cite{Zhang2022InterpretingPerturbations, moradi-samwald-2021-evaluating}. At the word level, we replace words with tokens, such as \verb|MASK|, as done in perturbation-based studies \cite{Madsen2021EvaluatingRetraining}. We also compare \verb|MASK| replacement with \verb|UNK| tokens replacement. %
We convert the entire word to l33t speak (\verb|l33t|) \cite{Eger2019TextSystems,Zhang2022InterpretingPerturbations}%
, and swap the word with a semantically related word (\verb|synonym|) using publicly available corpora \cite{Pavlick2015PPDBClassification, wordnet, nltk}, manually-made dictionaries (e.g., for public Twitter IDs) or randomly generated replacements (e.g., for URLs). Not all words have valid synonyms; therefore, we are only able to perturb about 16.2\% of words in the Hummingbird dataset and 18.4\% of the SemEval dataset. These mainly consist of rare or slang words, and non-parseable hashtags or misspellings in the case of the SemEval dataset. Our precise rules for synonym replacement can be found in Appendix \ref{sec:APP_synonyms}.

\begin{figure*}
    \centering
    \begin{adjustbox}{max width=1.0\textwidth,center}
    \includegraphics[scale=0.37]{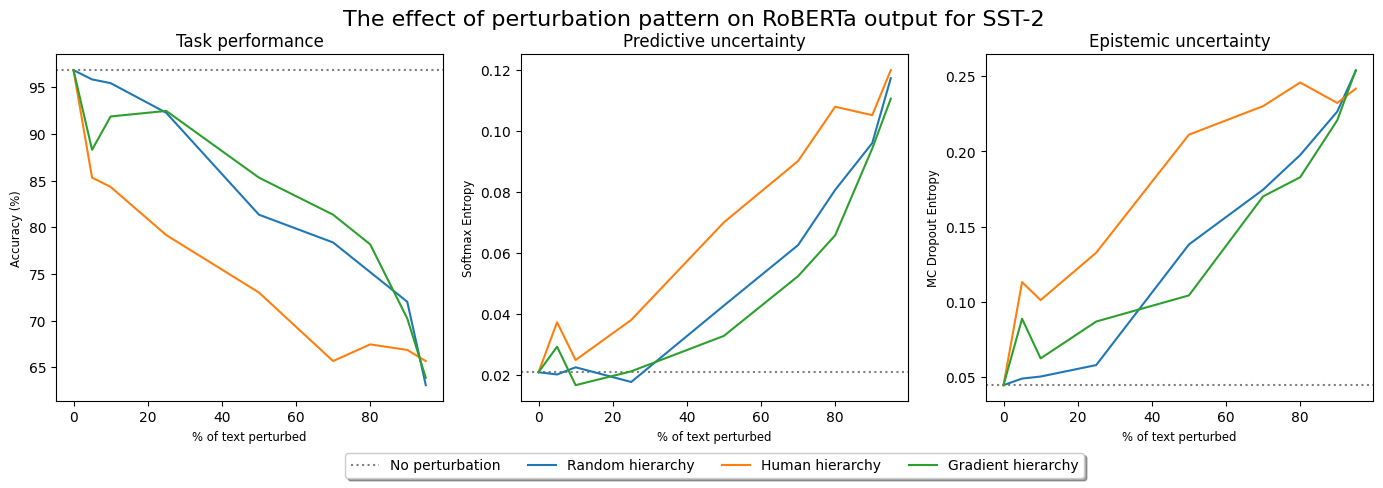}
    \end{adjustbox}
    \caption{The effect of increasing text perturbation (averaged across perturbation type) on RoBERTa performance and uncertainty across three different hierarchies: (1) Random; (2) Human, following human annotation and POS tags; and (3) Gradient, following ranking of Hotflip gradients. Dotted lines show the value at $\alpha=0.0$.}
    \label{fig:RQ1pattern}
\end{figure*}

\subsection{Explanation techniques}
 We focus on local gradient-based explanations, which use backpropagation to compute a saliency heatmap over input features for a specific datapoint to audit a model's decision. These explanations have been shown to perform best across many metrics, models, and tasks \cite{Atanasova2020AClassification}, and, compared to perturbation-based techniques, like LIME, which approximate model performance, are model-centric and should give a more faithful representation of the instability within the model, rather than the technique \cite{Zhang2019WhyExplanations}.
 The simplest implementation uses the gradient of the input as the saliency score \cite{Simonyan2013DeepMaps}; however, this can be very noisy \cite{Smilkov2017SmoothGrad:Noise}. Therefore, we rely on modified versions: \textbf{SmoothGrad} (\verb|SG|) returns the average saliency map obtained by perturbing the original input with Gaussian noise \cite{Smilkov2017SmoothGrad:Noise}. \textbf{Guided Backpropogation} (\verb|GBP|) uses a different computation of gradients (by ignoring all negative values) to visually improve its saliency maps \cite{Springenberg2014StrivingNet}. \textbf{InputXGradients} (\verb|IXG|) considers both the importance of the feature and the strength of the expressed dimension \cite{Shrikumar2016NotDifferences}. \textbf{IntegratedGradients} (\verb|IG|) accumulates the gradients between an input of interest and a neutral baseline \cite{Sundararajan2017AxiomaticNetworks}. We use the Captum implementations \cite{captum2019github}.

\subsection{Evaluation design}
\label{sec:eval}
\color{black}
Throughout our investigations, we use the following metrics: To measure model performance, we use \textbf{accuracy}. To measure uncertainty, we use two measures: Following similar perturbation studies, we include \textbf{predictive uncertainty}, the conflated, popular measure of uncertainty, measured via the entropy of the softmax logits (to reduce overconfidence \cite{Pearce2021UnderstandingUncertainty}). We define \textbf{epistemic uncertainty}, the uncertainty in the model's parameters, using MC Dropout entropy, following \citet{Kendall2016WhatVision}. We define \textbf{explanation robustness} as the average Pearson correlation between saliency map\textsubscript{$\alpha=.05$} and saliency map\textsubscript{$\alpha=.00$}. %
We also define \textbf{explanation plausibility} as the Mean Average Precision (MAP) of model gradients to the human annotations. There are many metrics to evaluate explanation quality, and each with pitfalls \cite{ju-etal-2022-logic}; we chose this one for its applicability for human-XAI collaboration and evaluation. We first evaluate its suitability by assessing the change in model performance and confidence between the perturbation of human- and gradient-ranked salient tokens. For all saliency map comparisons, we combine all gradients back to word level%
.

We first look at general trends in performance, uncertainty, and explanation plausibility with increasing perturbation across models and datasets. As a Kolmogorov–Smirnov test of the plausibility and uncertainty measures violates the assumption of normality ($p<10^{-5}$), we use Spearman's Rank Correlation (SciPy v1.11.4) to find the correlation between the explanation plausibility and uncertainty measures at a datapoint level for correctly-predicted %
datapoints. We then compare the robustness of the saliency maps across model, dataset, and perturbation type.

\color{black}

  \begin{figure*}
    \centering
    \begin{adjustbox}{max width=1.0\textwidth,center}
    \includegraphics[scale=0.37]{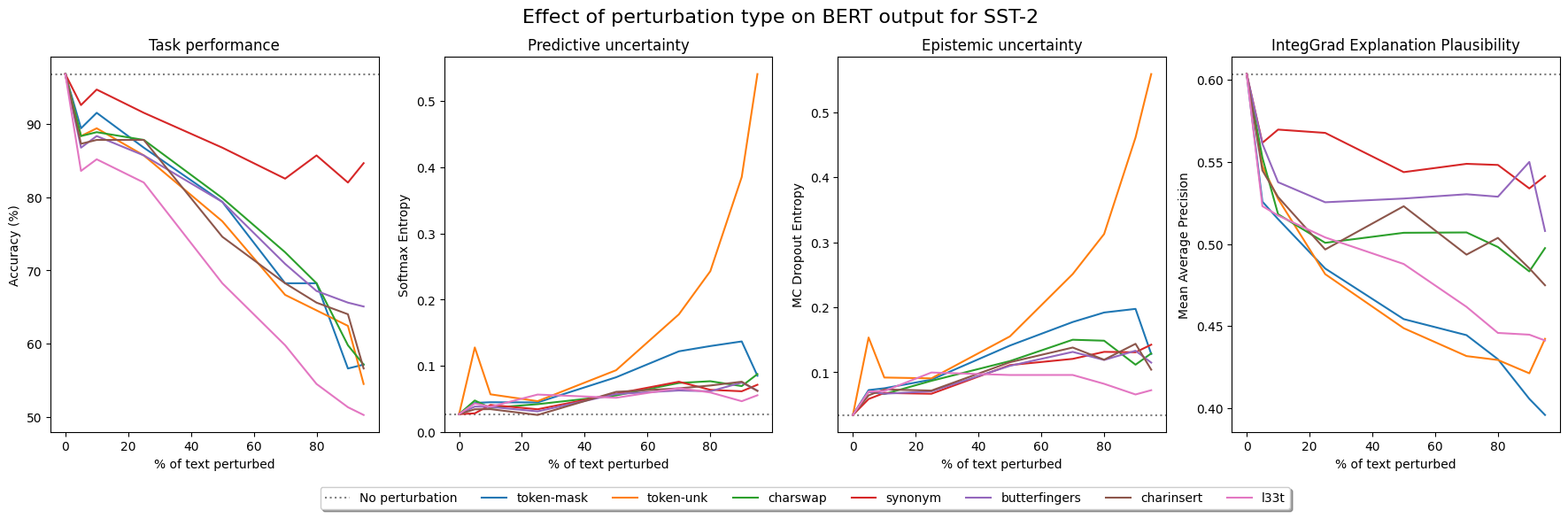} %
    \end{adjustbox}
    \caption{The effect of increasing text perturbation on BERT accuracy, uncertainty and explanation plausibility across the different types of perturbation, averaged across perturbation hierarchy. Dotted lines show $\alpha=0.0$.}
    \label{fig:RQ1nk}
\end{figure*}

\section{Results}
\label{sec:res_all}
We present the motivation and results of each investigation. We show the results for SST-2 but summarize and interpret the results for all investigated datasets; the data for all datasets are in the Appendix.
\subsection{Noise on uncertainty and explanations} 
\label{sec:RQ1_res}

\paragraph{The effect of perturbation hierarchy}
To ensure the faithfulness of using human annotations as ground-truth for salient tokens \cite{ju-etal-2022-logic}, we showcase the impact of different hierarchies of perturbation (as described in \S\ref{sec:perturb}) on model performance, uncertainty and explanations in Figure \ref{fig:RQ1pattern} and Appendix \ref{app:RQ1-1_all}. All perturbations impair model performance, uncertainty, and explanation plausibility, but human-hierarchical perturbation has the greatest impact up to very high levels of perturbation across all tasks, suggesting that these human-salient tokens are vital signals for the models. While random and gradient-based perturbation generally have similar impacts on task performance, uncertainty and explanation plausibility, gradient-based perturbation strategies have a stronger impact on these metrics at low levels of perturbation ($\alpha=.05$), which lessens with slightly more perturbation ($\alpha=.1$), and suggests that gradient-based perturbation techniques have their greatest efficacy at low levels of perturbation.

\paragraph{The effect of perturbation type}

\color{black} 
To assess the impact of our various perturbation types, we show the effect of the investigated noise types (see Table \ref{tab:noises}) in Figure \ref{fig:RQ1nk}. Though all perturbation types adversely impact task performance and explanation plausibility, this effect is typically smaller for more `realistic' perturbations, especially \verb|synonym| and \verb|butterfingers|. Across datasets and models, we typically see that special token replacements have the greatest detrimental effect (particularly \verb|MASK|). This trend can be seen across models and datasets, as shown in Appendix \ref{app:RQ1-2_all}. Typically, we see an inverse relationship between task performance and uncertainty for all perturbation types; In Figure \ref{fig:RQ1nk}, a steep decrease in performance from \verb|l33t| perturbation comes with little to no increase in either uncertainty measure. Furthermore, while we do often see a trend for uncertainty to increase with increased perturbation, it is not at the same rate of performance decline, mimicking over-confidence issues reported in other studies \cite{Feng2018PathologiesDifficult,Gupta2021BERTUnderstanding, Pearce2021UnderstandingUncertainty}. As we show in Appendix \ref{app:RQ1-2_all}, this is especially pronounced with RoBERTa.

\color{black}

\subsection{The relationship between uncertainty and explanation plausibility}
\label{sec:RQ2_res}

\color{black}
\begin{table*}[!htbp]
\small
\centering
\begin{adjustbox}{max width=1.0\textwidth,center}
\begin{tblr}{
  row{3} = {r},
  row{4} = {r},
  row{5} = {r},
  row{6} = {r},
  row{7} = {r},
  row{8} = {r},
  row{9} = {r},
  row{10} = {r},
  row{11} = {r},
  row{12} = {r},
  row{13} = {r},
  row{14} = {r},
  row{15} = {r},
  row{16} = {r},
  row{17} = {r},
  row{18} = {r},
  cell{1}{3} = {c=8}{c},
  cell{1}{11} = {c=8}{c},
  cell{2}{3} = {c=4}{c},
  cell{2}{7} = {c=4}{c},
  cell{2}{11} = {c=4}{c},
  cell{2}{15} = {c=4}{c},
  cell{4}{1} = {r=5}{},
  cell{9}{1} = {r=5}{},
  cell{14}{1} = {r=5}{},
  vline{4} = {1}{},
  vline{4,8,12} = {2}{},
  vline{3,7,11,15} = {3-18}{},
  hline{1,19} = {-}{0.08em},
  hline{2-4} = {-}{},
  hline{9,14} = {2-18}{},
  columns = {colsep=3pt}
}
                    &                & \textbf{Before Perturbation}    &                 &                &                 &                                &                 &                &                & \textbf{Including Perturbed Text} &                 &                &                &                                &                 &                 &                \\
                    &                & \textbf{Predictive uncertainty} &                 &                &                 & \textbf{Epistemic uncertainty} &                 &                &                & \textbf{Predictive uncertainty}   &                 &                &                & \textbf{Epistemic uncertainty} &                 &                 &                \\
\textbf{dataset}    & \textbf{model} & \textbf{GBP}                    & \textbf{IXG}    & \textbf{IG}    & \textbf{SG}     & \textbf{GBP}                   & \textbf{IXG}    & \textbf{IG}    & \textbf{SG}    & \textbf{GBP}                      & \textbf{IXG}    & \textbf{IG}    & \textbf{SG}    & \textbf{GBP}                   & \textbf{IXG}    & \textbf{IG}     & \textbf{SG}    \\
\textbf{SST-2}      & BERT           & 0.076                           & 0.068           & -0.128         & \textbf{-0.155} & -0.052                         & \textbf{-0.060} & 0.041          & 0.039          & \textbf{-0.104}                   & -0.099          & -0.069         & -0.069         & -0.240                         & -0.228          & \textbf{-0.248} & -0.219         \\
                    & ELECTRA        & 0.040                           & 0.002           & -0.050         & \textbf{-0.089} & \textbf{-0.127}                & -0.065          & -0.050         & -0.058         & \textbf{-0.096}                   & \textbf{-0.096} & -0.043         & -0.050         & \textbf{-0.383}                & -0.380          & -0.164          & -0.175         \\
                    & RoBERTa        & \textbf{0.088}                  & 0.048           & 0.030          & -0.000          & \textbf{-0.367}                & -0.330          & -0.174         & -0.200         & \textbf{-0.124}                   & -0.101          & -0.084         & -0.069         & \textbf{-0.357}                & -0.324          & -0.267          & -0.246         \\
                    & GPT2           & 0.078                           & -0.033          & \textbf{0.124} & -0.014          & -0.150                         & \textbf{-0.237} & -0.036         & -0.088         & \textbf{-0.092}                   & -0.068          & -0.013         & -0.004         & -0.232                         & \textbf{-0.241} & -0.094          & -0.068         \\
                    & OPT            & \textbf{-0.205}                 & \textbf{-0.205} & -0.030         & -0.030          & \textbf{-0.159}                & \textbf{-0.159} & -0.099         & -0.099         & \textbf{-0.152}                   & \textbf{-0.152} & -0.130         & -0.130         & \textbf{-0.219}                & \textbf{-0.219} & -0.109          & -0.109         \\
\textbf{SemEval}    & BERT           & 0.237                           & 0.248           & 0.238          & \textbf{0.249}  & 0.235                          & \textbf{0.247}  & 0.234          & \textbf{0.247} & 0.149                             & \textbf{0.165}  & 0.150          & \textbf{0.165} & 0.151                          & \textbf{0.166}  & 0.148           & 0.164          \\
                    & ELECTRA        & 0.200                           & \textbf{0.232}  & 0.199          & \textbf{0.232}  & 0.201                          & \textbf{0.233}  & 0.199          & 0.231          & 0.162                             & \textbf{0.169}  & 0.162          & \textbf{0.169} & 0.163                          & \textbf{0.171}  & 0.162           & 0.170          \\
                    & RoBERTa        & 0.213                           & \textbf{0.234}  & 0.212          & \textbf{0.234}  & 0.215                          & \textbf{0.235}  & 0.213          & \textbf{0.235} & 0.149                             & \textbf{0.155}  & 0.148          & 0.154          & 0.149                          & \textbf{0.155}  & 0.147           & 0.153          \\
                    & GPT2           & \textbf{0.220}                  & 0.181           & 0.218          & 0.182           & \textbf{0.221}                 & 0.184           & 0.219          & 0.181          & \textbf{0.127}                    & 0.120           & \textbf{0.127} & 0.121          & \textbf{0.128}                 & 0.122           & 0.127           & 0.120          \\
                    & OPT            & 0.224                           & 0.224           & \textbf{0.226} & \textbf{0.226}  & \textbf{0.230}                 & \textbf{0.230}  & 0.224          & 0.224          & 0.164~                            & 0.164           & \textbf{0.165} & \textbf{0.165} & \textbf{0.167}                 & \textbf{0.167}  & 0.164           & 0.164          \\
\textbf{HateXplain} & BERT           & 0.268                           & \textbf{0.270}  & 0.265          & 0.262           & 0.211                          & 0.229           & 0.263          & \textbf{0.267} & 0.293                             & 0.178           & \textbf{0.297} & 0.181          & 0.243                          & 0.139           & \textbf{0.259}  & 0.148          \\
                    & ELECTRA        & 0.565                           & 0.458           & \textbf{0.573} & 0.464           & 0.539                          & 0.430           & \textbf{0.568} & 0.462          & 0.444                             & 0.240           & \textbf{0.452} & 0.247          & 0.425                          & 0.221           & \textbf{0.448}  & 0.244          \\
                    & RoBERTa        & \textbf{0.529}                  & 0.434           & 0.517          & 0.424           & 0.502                          & 0.407           & \textbf{0.503} & 0.408          & \textbf{0.396}                    & 0.218           & 0.390          & 0.213          & 0.371                          & 0.195           & \textbf{0.379}  & 0.201          \\
                    & GPT2           & \textbf{0.393}                  & 0.278           & 0.386          & 0.278           & 0.380                          & 0.270           & \textbf{0.399} & 0.284          & \textbf{0.300}                    & 0.106           & 0.298          & 0.105          & 0.291                          & 0.097           & \textbf{0.304}  & 0.110          \\
                    & OPT            & \textbf{0.459}                  & \textbf{0.459}  & 0.428          & 0.428           & 0.432                          & 0.432           & \textbf{0.456} & \textbf{0.456} & \textbf{0.408}                    & \textbf{0.408}  & 0.391          & 0.391          & 0.382                          & 0.382           & \textbf{0.410} & \textbf{0.410} 
\end{tblr}
\end{adjustbox}
\caption{The Spearman Rank Correlation between explanation plausibility and both measures of uncertainty across model, dataset, and saliency technique. We bold the strongest correlation for each comparison.}
\label{tab:unc_rel}
\end{table*}

\color{black}
While we see a general increase in uncertainty and a decrease in explanation quality with perturbation, we want to investigate at a data point level if greater uncertainty of a model output implies lower plausibility across explanation techniques. Therefore, we assess the correlation between uncertainty and explanation plausibility across all datasets, saliency maps, and models in Table \ref{tab:unc_rel}. \color{black} We see similar patterns in correlation between all attribution methods and before and after perturbation with some exceptions: SmoothGrad (\verb|SG|) typically shows much weaker correlation after perturbation, whereas Guided Backpropagation (\verb|GBP|) and Integrated Gradients (\verb|IG|) show the strongest. While we would expect increased uncertainty to imply decreased explanation plausibility, this is not always the case; While SST-2 shows a low negative correlation between epistemic uncertainty and explanation plausibility after perturbation, this is not the case for SemEval and HateXplain, which has a low-to-moderate positive correlation that exists before and after perturbation. 
\color{black}
Therefore, greater uncertainty of an output does not necessarily imply a degradation in explanation quality. %

\color{black}

\subsection{Robustness across perturbation type}
\label{sec:RQ3_res}

\begin{figure*}
    \centering
    \begin{adjustbox}{max width=1.0\textwidth,center}
    \includegraphics[scale=0.32]{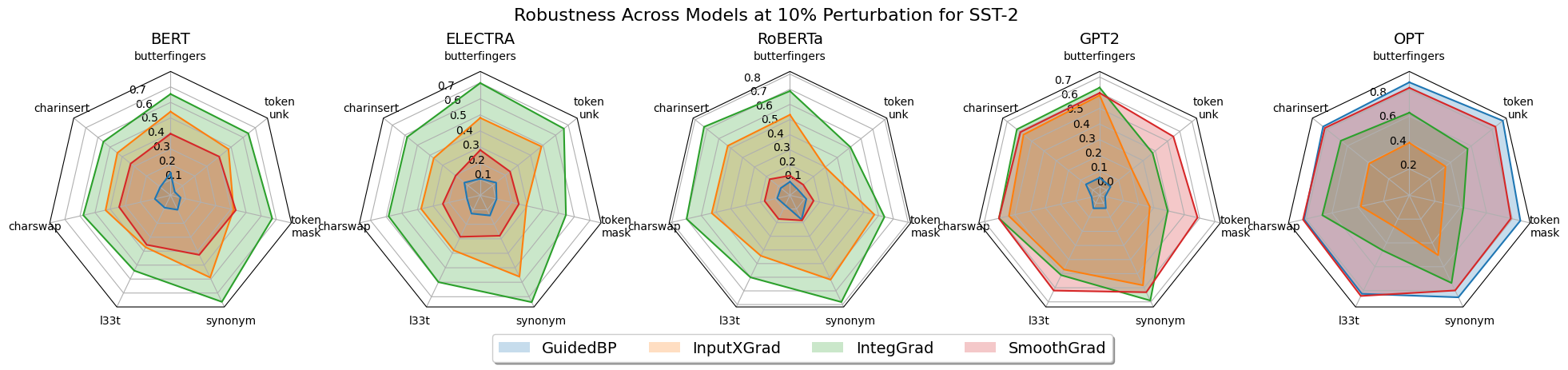}
    \end{adjustbox}
    \caption{Model-level differences of the correlation to the unperturbed saliency map at low levels of perturbation. We separately show the effect on BERT, RoBERTa, ELECTRA, GPT2 and OPT.}
    \label{fig:RQ3_1mod}
\end{figure*}

\color{black}
To investigate how noise introduces instability in explanation maps, we assess how saliency maps are impacted by our perturbations. We show the robustness values across each model in Figure \ref{fig:RQ3_1mod} and see distinct patterns that are shared across most saliency maps. For example, Integrated Gradients (\verb|IG|), InputXGrad (\verb|IXG|), and Guided Backpropagation (\verb|GBP|) all show reduced robustness to \verb|l33t| perturbations and increased robustness to \verb|synonym| at low-levels of perturbation. This aligns with the differences we see in task performance in Figure \ref{fig:RQ1nk}. We also see different patterns emerge across different models; RoBERTa has general lower robustness to \verb|UNK| tokens, and ELECTRA to \verb|MASK|. \verb|IG| shows the greatest overall robustness for the models BERT, RoBERTa, and ELECTRA, but SmoothGrad (\verb|SG|) has greater robustness for GPT2 and OPT. \verb|GBP|, which typically has low robustness for all other models, shows the greatest robustness for OPT. These patterns are preserved across datasets (See Appendix \ref{app:RQ3_all}), where perturbations to which we find decreased robustness typically also further deteriorate model performance. Furthermore, on datasets with lower performance (HateXplain), we also see decreased overall robustness.
\color{black}

\paragraph{In summary} While perturbation decreases model performance and explanation plausibility, it has a task and perturbation-dependent effect on uncertainty. Furthermore, high uncertainty of an output does not necessarily imply low explanation plausibility, as we find a moderate positive correlation between the measures on some datasets.  %
\color{black} Where uncertainty measures fail to align with model accuracy on some perturbation patterns, saliency map robustness can provide additional indication of model performance patterns; Integrated Gradients typically shows the greatest robustness to all types of noise; however, we can see model-specific patterns in susceptibility to adversarial perturbation.

\color{black}
\section{Discussion}
In this section, we discuss the causes behind patterns seen across the experiments in \S\ref{sec:res_all}, which are further supplemented with extra analyses in our Appendix. We investigate over-arching patterns as well as patterns across perturbation types and datasets.

\paragraph{Overarching patterns} While noise consistently deteriorates model performance and explanation plausibility, the impact of increasing noise on model confidence varies across model and task. Unlike previous studies, we do not typically see an increase in confidence after perturbation \cite{Feng2018PathologiesDifficult,Gupta2021BERTUnderstanding}; though the observed decrease is not at the same rate of performance decline. However, both cited studies perturb at the word and sentence structure level, unlike our study. Furthermore, we see a similar pattern between the predictive and epistemic uncertainty measures, suggesting that over-confidence after perturbation stems from the training process. Overall, human-based perturbations have the strongest effect on task performance and uncertainty measures, and %
gradient-based perturbation is only more effective than random perturbation at low levels of noise ($\alpha=.05$). This suggests that the human-generated annotations of each dataset are faithful indicators of true saliency, as their perturbation degrades model performance more than gradient-based approaches, further justifying our use of plausibility as a quality metric. 

\color{black}

\paragraph{Perturbation-level patterns} Across all models, realistic perturbations, such as \verb|butterfingers| or \verb|synonym| have the smallest impact on task performance and explanation plausibility, yet masking has the greatest impact. Furthermore, \verb|MASK| has the greatest effect on both measures of uncertainty. We expect that the embedding layer of the PLMs is better equipped to handle synonym-level perturbations, allowing the hidden representations of the input to change minimally, so that model performance and explanations are minimally impacted. %
Furthermore, we explore model-level differences in Appendix \ref{app:RQ1-2_all} and \ref{app:RQ3_all} and find that the perturbations that have the maximal impact on model performance (e.g. \verb|MASK| for ELECTRA in SemEval) also uniquely impact saliency maps at low levels of perturbation. This suggests that the instability we see in explanations can provide some signal to model performance in the absence of labels. While SmoothGrad shows good all-around robustness to noise due to its regularization, it does not show specific patterns in model instability. In contrast, Integrated Gradients has relatively high robustness for smaller language models at low levels of perturbation and shows increased robustness to perturbations that minimally impact model performance (\verb|synonym| and \verb|butterfinger|). While lack of robustness is typically viewed as a deficiency of an XAI technique \cite{Hedstrom2023Quantus:Beyond}, we believe it can also be a signal of model instability: Epistemic uncertainty approximation measures work to perturb the model decision boundary to obtain a datapoint's likelihood of class correspondence as an indicator of uncertainty. In contrast, perturbation, by introducing stochasticity in an input (much like SmoothGrad), also suggests the proximity of a datapoint to the decision boundary. In cases where popular uncertainty measures are prone to over-confidence, a lack of robustness may give some indication of uncertainty for a particular data point.

\color{black}
\paragraph{Dataset-level patterns} The relationship between uncertainty and explanation plausibility after perturbation varies across datasets. For HateXplain, \verb|UNK| and \verb|l33t| surprisingly reduce uncertainty (see Appendix \ref{app:RQ1-2_all}); this could explain the positive correlation between uncertainty and explanation plausibility for the dataset, as highly perturbed examples will show lower plausibility, yet lower uncertainty. The dataset is compiled from Twitter, and character substitutions may hide potentially offensive terms. While we do not see a significant class difference regarding the proportion of words containing letters and numbers (0:0.695\%, 1:0.975\%, 2:0.912\%), at manual inspection, we find examples of l33t-like speak in Classes 0 and 2 (e.g. `h0e') that we do not find in the neutral class (e.g. `WW2'). The existence of these examples in the training data may have made the noise an indicator of a class, owing to the high ``learnability'' of this perturbation \cite{Zhang2022InterpretingPerturbations}, creating the positive correlation between uncertainty, output quality and explanation plausibility.\\ In Appendix  \ref{app:noise_corr}, we investigate if particular perturbation types are salient and find that saliency is not attributed to l33t noise, suggesting it is not a class indicator. Furthermore, we also see a weaker, positive relationship with the Twitter-based SemEval dataset. Therefore, though one would expect to see an inverse relationship between uncertainty and explanation plausibility, we only see this behaviour with the SST-2 dataset; we posit that models trained with noisy data instead show a positive relationship between uncertainty and explanation plausibility. When these models express greater uncertainty, they are more precise at identifying salient tokens, adding to other reported performance improvements after training models with noisy data \cite{Yu2024ExploringSystems}. We show in Appendix \ref{sec:unc_high} that, at very high perturbation, the strength of this relationship weakens (due to lack of meaningful tokens), but can remain weakly positive for simple tasks. %

\color{black}
\paragraph{In summary} The results suggest that the effect of perturbation on language models must be considered holistically across noise type and training data; realistic perturbations, like synonyms and mispellings, which are expected to be more prevalent in social media, have a smaller impact on performance, uncertainty, and explanations. While uncertainty is not always a faithful indicator of local instability, weakened robustness to perturbations can provide additional information for model performance. Furthermore, as high uncertainty does not necessarily imply low explanation plausibility with noisy datasets, noise-augmented training may not only help model performance in out-of-domain tasks, but may also help ensure coherent explanations in low-confidence domains. Further research is required to devise a metric to estimate explanation quality at a datapoint-level, as current uncertainty measures are not reflective of explanation quality. For future work, we recommend the use of Integrated Gradients for smaller language models as it gives a more holistic depiction of model performance in adversarial conditions; however, as models scale in size, other gradient-based explanation techniques are more robust. %

\color{black}
\section{Conclusion}
We provide an empirical investigation across language models, noise perturbations, and saliency maps to investigate a relationship between uncertainty and explanation plausibility. Following an array of perturbation techniques, we show that noise injection simultaneously affects model performance, uncertainty, and explanation plausibility. %
\color{black}
We do not find a strong negative relationship between uncertainty and explanation plausibility; model fine-tuned with noisy data typically show a moderately positive correlation between plausibility and uncertainty, which suggests that these models may even be better at identifying salient tokens when uncertain. We also show that the instability of a saliency map to noise can also provide insights into a model's performance, and suggest Integrated Gradients for future work in Human-XAI collaboration, due to its robustness to noise for smaller language models. %
\color{black}

\section*{Limitations}

We do not investigate aleatoric uncertainty in this study, as our experimental setup intended to simulate epistemic uncertainty by introducing noise not present in the training data. However, we do assess across different dataset sources, with differing levels of latent noise in the data, and, therefore, differing aleatoric uncertainty, and find highly correlated results for a shared task. Future work should consider further disambiguating aleatoric uncertainty in their comparisons. In addition, given our investigation into epistemic uncertainty, it could be interesting to assess how the observed robustness changes in models fine-tuned with noise-augmented training data. Future studies could consider simulating uncertainty in other methods, perhaps at other points of the experimental pipeline.

Though we do compare many popular language models, more model types would have made an interesting comparison. Models with visual encoding, for example PIXEL \cite{Rust2023LANGUAGEPIXELS}, may handle different types of noise differently; visual perturbations, like l33t speak, may show a lesser effect on PIXEL model performance and confidence, whereas semantic changes, like synonym replacement, may have a larger effect. However, given the format of our study, the saliency maps would be difficult to compare across all model types. It would also have been interesting to explore larger language models ($>1$B parameters), like LLAMA \cite{Touvron2023LLaMA:Models}; however, our focus on gradient-based explanations makes such an investigation very computationally expensive. Furthermore, our requirement for human annotations limited the possible number of datasets for investigation; however, our pilot studies on other popular NLP tasks found very similar results to those reported in this study.

\color{black}

\section*{Acknowledgements}

$\begin{array}{l}\includegraphics[width=1cm]{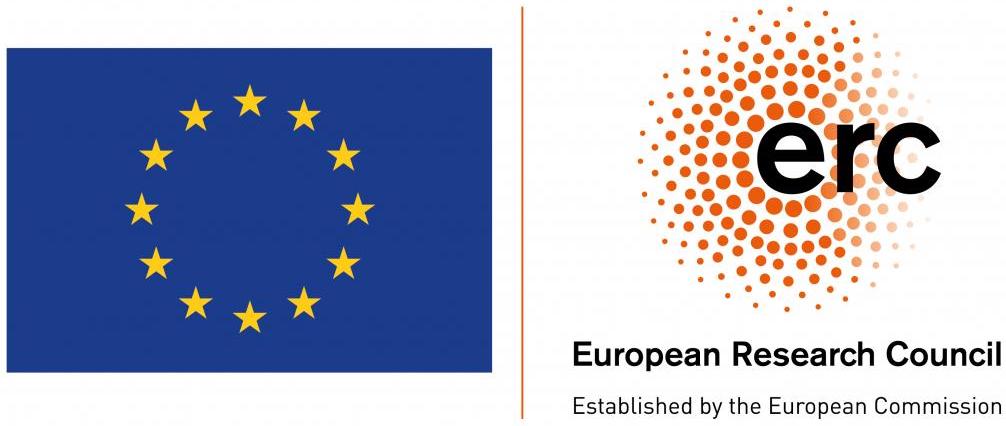} \end{array}$ 
This research was co-funded by the European Union (ERC, ExplainYourself, 101077481), and supported by the Pioneer Centre for AI, DNRF grant number P1. We are also funded by the Villum and Velux Foundations Algorithms, Data and Democracy (ADD) grant, as well as the ERRATUM UCPH Data+ grant. Views and opinions expressed are however those of the author(s) only and do not necessarily reflect those of the European Union or the European Research Council. Neither the European Union nor the granting authority can be held responsible for them. We thank the anonymous reviewers for their helpful suggestions.

\bibliography{anthology,ref,tools,references}

\begin{thebibliography}{61}
\expandafter\ifx\csname natexlab\endcsname\relax\def\natexlab#1{#1}\fi

\bibitem[{Adebayo et~al.(2018)Adebayo, Gilmer, Muelly, Goodfellow, Hardt, and
  Kim}]{Adebayo2018SanityMaps}
Julius Adebayo, Justin Gilmer, Michael Muelly, Ian Goodfellow, Moritz Hardt,
  and Been Kim. 2018.
\newblock \href {https://arxiv.org/abs/1810.03292} {{Sanity Checks for Saliency
  Maps}}.

\bibitem[{Alipanahi et~al.(2022)Alipanahi, Hormozdiari, D'amour, Heller,
  Moldovan, Adlam, Beutel, Chen, Deaton, Eisenstein, Hoffman, Houlsby, Hou,
  Jerfel, Karthikesalingam, Lucic, Ma, Mclean, Mincu, Mitani, Montanari, Nado,
  Natarajan, Nielson, Osborne, Raman, Ramasamy, Sayres, Schrouff, Seneviratne,
  Sequeira, Suresh, Veitch, Vladymyrov, Wang, Webster, Yadlowsky, Yun, Zhai,
  and Sculley}]{Alipanahi2022UnderspecificationLearning}
Babak Alipanahi, Farhad Hormozdiari, Alexander D'amour, Katherine Heller, Dan
  Moldovan, Ben Adlam, Alex Beutel, Christina Chen, Jonathan Deaton, Jacob
  Eisenstein, Matthew~D Hoffman, Neil Houlsby, Shaobo Hou, Ghassen Jerfel, Alan
  Karthikesalingam, Mario Lucic, Yian Ma, Cory Mclean, Diana Mincu, Akinori
  Mitani, Andrea Montanari, Zachary Nado, Vivek Natarajan, Christopher Nielson,
  Thomas~F Osborne, Rajiv Raman, Kim Ramasamy, Rory Sayres, Jessica Schrouff,
  Martin Seneviratne, Shannon Sequeira, Harini Suresh, Victor Veitch, Max
  Vladymyrov, Xuezhi Wang, Kellie Webster, Steve Yadlowsky, Taedong Yun,
  Xiaohua Zhai, and D~Sculley. 2022.
\newblock \href {http://jmlr.org/papers/v23/20-1335.html.} {{Underspecification
  Presents Challenges for Credibility in Modern Machine Learning}}.
\newblock 23:1--61.

\bibitem[{Alvarez-Melis and Jaakkola(2018)}]{Alvarez-Melis2018OnMethods}
David Alvarez-Melis and Tommi~S. Jaakkola. 2018.
\newblock \href {https://arxiv.org/abs/1806.08049} {{On the Robustness of
  Interpretability Methods}}.

\bibitem[{Amorim et~al.(2023)Amorim, Abreu, Santos, Cortes, and
  Vila}]{Amorim2023EvaluatingPerturbations}
José~P. Amorim, Pedro~H. Abreu, João Santos, Marc Cortes, and Victor Vila.
  2023.
\newblock \href {https://doi.org/10.1016/j.ipm.2022.103225} {{Evaluating the
  faithfulness of saliency maps in explaining deep learning models using
  realistic perturbations}}.
\newblock \emph{Information Processing and Management}, 60(2).

\bibitem[{Atanasova et~al.(2020)Atanasova, Simonsen, Lioma, and
  Augenstein}]{Atanasova2020AClassification}
Pepa Atanasova, Jakob~Grue Simonsen, Christina Lioma, and Isabelle Augenstein.
  2020.
\newblock \href {https://arxiv.org/abs/2009.13295} {{A Diagnostic Study of
  Explainability Techniques for Text Classification}}.

\bibitem[{Augenstein et~al.(2023)Augenstein, Baldwin, Cha, Chakraborty,
  Ciampaglia, Corney, DiResta, Ferrara, Hale, Halevy, Hovy, Ji, Menczer,
  Miguez, Nakov, Scheufele, Sharma, and Zagni}]{augenstein2023factuality}
Isabelle Augenstein, Timothy Baldwin, Meeyoung Cha, Tanmoy Chakraborty,
  Giovanni~Luca Ciampaglia, David Corney, Renee DiResta, Emilio Ferrara, Scott
  Hale, Alon Halevy, Eduard Hovy, Heng Ji, Filippo Menczer, Ruben Miguez,
  Preslav Nakov, Dietram Scheufele, Shivam Sharma, and Giovanni Zagni. 2023.
\newblock \href {http://arxiv.org/abs/2310.05189} {{Factuality Challenges in
  the Era of Large Language Models}}.

\bibitem[{Bang et~al.(2023)Bang, Cahyawijaya, Lee, Dai, Su, Wilie, Lovenia, Ji,
  Yu, Chung, Do, Xu, and Fung}]{bang2023multichatgpt}
Yejin Bang, Samuel Cahyawijaya, Nayeon Lee, Wenliang Dai, Dan Su, Bryan Wilie,
  Holy Lovenia, Ziwei Ji, Tiezheng Yu, Willy Chung, Quyet~V. Do, Yan Xu, and
  Pascale Fung. 2023.
\newblock \href {https://doi.org/10.48550/arXiv.2302.04023} {{A Multitask,
  Multilingual, Multimodal Evaluation of ChatGPT on Reasoning, Hallucination,
  and Interactivity}}.
\newblock \emph{CoRR}, abs/2302.04023.

\bibitem[{Bauer et~al.(2023)Bauer, Moritz Von~Zahn, Hinz, and
  Von~Zahn}]{Bauer2023PleaseKnowledge}
Kevin Bauer, |~Moritz Von~Zahn, Oliver Hinz, and Moritz Von~Zahn. 2023.
\newblock {Please Take Over: XAI, Delegation of Authority, and Domain
  Knowledge}.

\bibitem[{Brown and Talbert(2022)}]{Brown2022UsingUncertainty}
Katherine~E Brown and Douglas~A Talbert. 2022.
\newblock {Using Explainable AI to Measure Feature Contribution to
  Uncertainty}.

\bibitem[{Bykov et~al.(2020)Bykov, H{\"{o}}hne, M{\"{u}}ller, Nakajima, and
  Kloft}]{Bykov2020HowNetworks}
Kirill Bykov, Marina M.~C. H{\"{o}}hne, Klaus-Robert M{\"{u}}ller, Shinichi
  Nakajima, and Marius Kloft. 2020.
\newblock \href {http://arxiv.org/abs/2006.09000} {{How Much Can I Trust You?
  -- Quantifying Uncertainties in Explaining Neural Networks}}.

\bibitem[{Chai(2018)}]{Chai2018UncertaintyInterpretability}
Lucy~R Chai. 2018.
\newblock {Uncertainty Estimation in Bayesian Neural Networks And Links to
  Interpretability}.

\bibitem[{Chen et~al.(2022)Chen, Subhash, Havasi, Pan, Doshi-Velez, and
  Paulson}]{Chen2022WHATEXPLANATIONS}
Zixi Chen, Varshini Subhash, Marton Havasi, Weiwei Pan, Finale Doshi-Velez, and
  John~A Paulson. 2022.
\newblock {What Makes A Good Explanation?: A Harmonized View Of Properties Of
  Explanations}.

\bibitem[{Chromik et~al.(2021)Chromik, Eiband, Buchner, Kr{\"{u}}ger, and
  Butz}]{Chromik2021IAI}
Michael Chromik, Malin Eiband, Felicitas Buchner, Adrian Kr{\"{u}}ger, and
  Andreas Butz. 2021.
\newblock \href {https://doi.org/10.1145/3397481.3450644} {{I Think i Get Your
  Point, AI! The Illusion of Explanatory Depth in Explainable AI}}.
\newblock In \emph{International Conference on Intelligent User Interfaces,
  Proceedings IUI}, pages 307--317. Association for Computing Machinery.

\bibitem[{Clark et~al.(2020)Clark, Luong, Brain, Le~Google~Brain, and
  Manning}]{Clark2020ELECTRA:GENERATORS}
Kevin Clark, Minh-Thang Luong, Google Brain, Quoc~V Le~Google~Brain, and
  Christopher~D Manning. 2020.
\newblock \href {https://github.com/google-research/} {{ELECTRA: PRE-TRAINING
  TEXT ENCODERS AS DISCRIMINATORS RATHER THAN GENERATORS}}.

\bibitem[{Devlin et~al.(2018)Devlin, Chang, Lee, and
  Toutanova}]{Devlin2018BERT:Understanding}
Jacob Devlin, Ming-Wei Chang, Kenton Lee, and Kristina Toutanova. 2018.
\newblock \href {http://arxiv.org/abs/1810.04805} {{BERT: Pre-training of Deep
  Bidirectional Transformers for Language Understanding}}.

\bibitem[{Ebrahimi et~al.(2018)Ebrahimi, Rao, Lowd, and
  Dou}]{Ebrahimi2018HotFlip:Classification}
Javid Ebrahimi, Anyi Rao, Daniel Lowd, and Dejing Dou. 2018.
\newblock \href {https://www.di.unipi.it/} {{HotFlip: White-Box Adversarial
  Examples for Text Classification}}.
\newblock pages 31--36.

\bibitem[{Eger et~al.(2019)Eger, G{\"{u}}l~{\c{S}}ahin, R{\"{u}}ckl{\'{e}},
  Lee, Schulz, Mesgar, Swarnkar, and Simpson}]{Eger2019TextSystems}
Steffen Eger, Gözde G{\"{u}}l~{\c{S}}ahin, Andreas R{\"{u}}ckl{\'{e}}, Ji-Ung
  Lee, Claudia Schulz, Mohsen Mesgar, Krishnkant Swarnkar, and Edwin Simpson.
  2019.
\newblock \href {www.aiphes.tu-darmstadt.de} {{Text Processing Like Humans Do:
  Visually Attacking and Shielding NLP Systems}}.
\newblock pages 1634--1647.

\bibitem[{Fellbaum(1998)}]{wordnet}
Christiane Fellbaum. 1998.
\newblock \href {https://mitpress.mit.edu/9780262561167/} {\emph{WordNet: An
  Electronic Lexical Database}}.
\newblock Bradford Books.

\bibitem[{Feng et~al.(2018)Feng, Wallace, Ii, Iyyer, Rodriguez, and
  Boyd-Graber}]{Feng2018PathologiesDifficult}
Shi Feng, Eric Wallace, Alvin~Grissom Ii, Mohit Iyyer, Pedro Rodriguez, and
  Jordan Boyd-Graber. 2018.
\newblock {Pathologies of Neural Models Make Interpretations Difficult}.
\newblock pages 3719--3728.

\bibitem[{Gal and Ghahramani(2015)}]{Gal2015DropoutLearning}
Yarin Gal and Zoubin Ghahramani. 2015.
\newblock \href {http://arxiv.org/abs/1506.02142} {{Dropout as a Bayesian
  Approximation: Representing Model Uncertainty in Deep Learning}}.

\bibitem[{Gupta et~al.(2021)Gupta, Kvernadze, and
  Srikumar}]{Gupta2021BERTUnderstanding}
Ashim Gupta, Giorgi Kvernadze, and Vivek Srikumar. 2021.
\newblock \href {http://arxiv.org/abs/2101.03453} {{BERT {\&} Family Eat Word
  Salad: Experiments with Text Understanding}}.

\bibitem[{Hayati et~al.(2021)Hayati, Kang, and Ungar}]{Hayati2021DoesLexica}
Shirley~Anugrah Hayati, Dongyeop Kang, and Lyle Ungar. 2021.
\newblock \href {http://arxiv.org/abs/2109.02738} {{Does BERT Learn as Humans
  Perceive? Understanding Linguistic Styles through Lexica}}.

\bibitem[{Hedstr{\"{o}}m et~al.(2023)Hedstr{\"{o}}m, Leander~Weber, Bareeva,
  Krakowczyk, Motzkus, Samek, Lapuschkin, and
  M-C~H{\"{o}}hne}]{Hedstrom2023Quantus:Beyond}
Anna Hedstr{\"{o}}m, tu-berlinde Leander~Weber, Dilyara Bareeva, Daniel
  Krakowczyk, Franz Motzkus, Wojciech Samek, Sebastian Lapuschkin, and Marina
  M-C~H{\"{o}}hne. 2023.
\newblock \href
  {https://github.com/understandable-machine-intelligence-lab/Quantus/}
  {{Quantus: An Explainable AI Toolkit for Responsible Evaluation of Neural
  Network Explanations and Beyond}}.
\newblock \emph{Journal of Machine Learning Research}, 24:1--11.

\bibitem[{Jin et~al.(2019)Jin, Jin, Zhou, and Szolovits}]{Jin2019IsEntailment}
Di~Jin, Zhijing Jin, Joey~Tianyi Zhou, and Peter Szolovits. 2019.
\newblock \href {http://arxiv.org/abs/1907.11932} {{Is BERT Really Robust? A
  Strong Baseline for Natural Language Attack on Text Classification and
  Entailment}}.

\bibitem[{Jin et~al.(2023)Jin, Li, and
  Hamarneh}]{Jin2023RethinkingPlausibility}
Weina Jin, Xiaoxiao Li, and Ghassan Hamarneh. 2023.
\newblock {Rethinking AI Explainability and Plausibility}.

\bibitem[{Ju et~al.(2022)Ju, Zhang, Yang, Jiang, Liu, and
  Zhao}]{ju-etal-2022-logic}
Yiming Ju, Yuanzhe Zhang, Zhao Yang, Zhongtao Jiang, Kang Liu, and Jun Zhao.
  2022.
\newblock \href {https://doi.org/10.18653/v1/2022.acl-long.407} {Logic traps in
  evaluating attribution scores}.
\newblock In \emph{Proceedings of the 60th Annual Meeting of the Association
  for Computational Linguistics (Volume 1: Long Papers)}, pages 5911--5922,
  Dublin, Ireland. Association for Computational Linguistics.

\bibitem[{Kendall and Gal(2016)}]{Kendall2016WhatVision}
Alex Kendall and Yarin Gal. 2016.
\newblock {What Uncertainties Do We Need in Bayesian Deep Learning for Computer
  Vision?}

\bibitem[{Kokhlikyan et~al.(2019)Kokhlikyan, Miglani, Martin, Wang, Reynolds,
  Melnikov, Lunova, and Reblitz-Richardson}]{captum2019github}
Narine Kokhlikyan, Vivek Miglani, Miguel Martin, Edward Wang, Jonathan
  Reynolds, Alexander Melnikov, Natalia Lunova, and Orion Reblitz-Richardson.
  2019.
\newblock Pytorch captum.
\newblock \url{https://github.com/pytorch/captum}.

\bibitem[{Lakkaraju and Bastani(2020)}]{Lakkaraju2020HowExplanations}
Himabindu Lakkaraju and Osbert Bastani. 2020.
\newblock \href {https://doi.org/10.1145/3375627.3375833} {{"How do I fool
  you?": Manipulating User Trust via Misleading Black Box Explanations}}.

\bibitem[{Liaw et~al.(2018)Liaw, Liang, Nishihara, Moritz, Gonzalez, and
  Stoica}]{liaw2018tune}
Richard Liaw, Eric Liang, Robert Nishihara, Philipp Moritz, Joseph~E Gonzalez,
  and Ion Stoica. 2018.
\newblock Tune: A research platform for distributed model selection and
  training.
\newblock \emph{arXiv preprint arXiv:1807.05118}.

\bibitem[{Liu et~al.(2019)Liu, Ott, Goyal, Du, Joshi, Chen, Levy, Lewis,
  Zettlemoyer, and Stoyanov}]{Liu2019RoBERTa:Approach}
Yinhan Liu, Myle Ott, Naman Goyal, Jingfei Du, Mandar Joshi, Danqi Chen, Omer
  Levy, Mike Lewis, Luke Zettlemoyer, and Veselin Stoyanov. 2019.
\newblock \href {http://arxiv.org/abs/1907.11692} {{RoBERTa: A Robustly
  Optimized BERT Pretraining Approach}}.

\bibitem[{Loper and Bird(2002)}]{nltk}
Edward Loper and Steven Bird. 2002.
\newblock \href {https://doi.org/10.48550/ARXIV.CS/0205028} {Nltk: The natural
  language toolkit}.

\bibitem[{Madsen et~al.(2021)Madsen, Meade, Adlakha, and
  Reddy}]{Madsen2021EvaluatingRetraining}
Andreas Madsen, Nicholas Meade, Vaibhav Adlakha, and Siva Reddy. 2021.
\newblock \href {https://arxiv.org/abs/2110.08412} {{Evaluating the
  Faithfulness of Importance Measures in NLP by Recursively Masking Allegedly
  Important Tokens and Retraining}}.

\bibitem[{Marx et~al.(2023)Marx, Park, Hasson, Wang, Ermon, and
  Huan}]{Marx2023ButAI}
Charlie Marx, Youngsuk Park, Hilaf Hasson, Yuyang Wang, Stefano Ermon, and Jun
  Huan. 2023.
\newblock {But Are You Sure? An Uncertainty-Aware Perspective on Explainable
  AI}.

\bibitem[{Mathew et~al.(2020)Mathew, Saha, Yimam, Biemann, Goyal, and
  Mukherjee}]{Mathew2020HateXplain:Detection}
Binny Mathew, Punyajoy Saha, Seid~Muhie Yimam, Chris Biemann, Pawan Goyal, and
  Animesh Mukherjee. 2020.
\newblock \href {http://arxiv.org/abs/2012.10289} {{HateXplain: A Benchmark
  Dataset for Explainable Hate Speech Detection}}.

\bibitem[{Moradi and Samwald(2021)}]{moradi-samwald-2021-evaluating}
Milad Moradi and Matthias Samwald. 2021.
\newblock \href {https://doi.org/10.18653/v1/2021.emnlp-main.117} {Evaluating
  the robustness of neural language models to input perturbations}.
\newblock In \emph{Proceedings of the 2021 Conference on Empirical Methods in
  Natural Language Processing}, pages 1558--1570, Online and Punta Cana,
  Dominican Republic. Association for Computational Linguistics.

\bibitem[{Nakov et~al.(2013)Nakov, Kozareva, Ritter, Rosenthal, Stoyanov, and
  Wilson}]{Nakov2013SemEval-2013Twitter}
Preslav Nakov, Zornitsa Kozareva, Alan Ritter, Sara Rosenthal, Veselin
  Stoyanov, and Theresa Wilson. 2013.
\newblock \href {http://wing.comp.nus.edu.sg/SMSCorpus/} {{SemEval-2013 Task 2:
  Sentiment Analysis in Twitter}}.
\newblock 2:312--320.

\bibitem[{Pavlick et~al.(2015)Pavlick, Rastogi, Ganitkevitch, Van~Durme, and
  Callison-Burch}]{Pavlick2015PPDBClassification}
Ellie Pavlick, Pushpendre Rastogi, Juri Ganitkevitch, Benjamin Van~Durme, and
  Chris Callison-Burch. 2015.
\newblock \href {https://code.google.com/p/word2vec/} {{PPDB 2.0: Better
  paraphrase ranking, fine-grained entailment relations, word embeddings, and
  style classification}}.
\newblock pages 425--430.

\bibitem[{Pearce et~al.(2021)Pearce, Brintrup, and
  Zhu}]{Pearce2021UnderstandingUncertainty}
Tim Pearce, Alexandra Brintrup, and Jun Zhu. 2021.
\newblock \href {http://arxiv.org/abs/2106.04972} {{Understanding Softmax
  Confidence and Uncertainty}}.

\bibitem[{Radford et~al.(2019)Radford, Wu, Child, Luan, Amodei, and
  Sutskever}]{Radford2019LanguageLearners}
Alec Radford, Jeffrey Wu, Rewon Child, David Luan, Dario Amodei, and Ilya
  Sutskever. 2019.
\newblock \href {https://github.com/codelucas/newspaper} {{Language Models are
  Unsupervised Multitask Learners}}.

\bibitem[{Ribeiro et~al.(2020)Ribeiro, Wu, Guestrin, and
  Singh}]{Ribeiro2020BeyondCheckList}
Marco~Tulio Ribeiro, Tongshuang Wu, Carlos Guestrin, and Sameer Singh. 2020.
\newblock \href {https://github.com/marcotcr/checklist} {{Beyond Accuracy:
  Behavioral Testing of NLP Models with CheckList}}.
\newblock pages 4902--4912.

\bibitem[{Rust et~al.(2023)Rust, Lotz, Bugliarello, Salesky, De~Lhoneux, and
  Elliott}]{Rust2023LANGUAGEPIXELS}
Phillip Rust, Jonas~F Lotz, Emanuele Bugliarello, Elizabeth Salesky, Miryam
  De~Lhoneux, and Desmond Elliott. 2023.
\newblock \href {https://github.com/xplip/pixel} {{LANGUAGE MODELLING WITH
  PIXELS}}.

\bibitem[{Schmidt et~al.(2020)Schmidt, Biessmann, and
  Teubner}]{Schmidt2020TransparencySystems}
Philipp Schmidt, Felix Biessmann, and Timm Teubner. 2020.
\newblock \href {https://doi.org/10.1080/12460125.2020.1819094} {{Transparency
  and trust in artificial intelligence systems}}.

\bibitem[{Shrikumar et~al.(2016)Shrikumar, Greenside, Shcherbina, and
  Kundaje}]{Shrikumar2016NotDifferences}
Avanti Shrikumar, Peyton Greenside, Anna Shcherbina, and Anshul Kundaje. 2016.
\newblock \href {http://arxiv.org/abs/1605.01713} {{Not Just a Black Box:
  Learning Important Features Through Propagating Activation Differences}}.

\bibitem[{Simonyan et~al.(2013)Simonyan, Vedaldi, and
  Zisserman}]{Simonyan2013DeepMaps}
Karen Simonyan, Andrea Vedaldi, and Andrew Zisserman. 2013.
\newblock \href {http://arxiv.org/abs/1312.6034} {{Deep Inside Convolutional
  Networks: Visualising Image Classification Models and Saliency Maps}}.

\bibitem[{Sinha et~al.(2021)Sinha, Parthasarathi, Pineau, and
  Williams}]{Sinha2021UnNaturalInference}
Koustuv Sinha, Prasanna Parthasarathi, Joelle Pineau, and Adina Williams. 2021.
\newblock \href {https://github.com/facebookresearch/unlu.} {{UnNatural
  Language Inference}}.
\newblock pages 7329--7346.

\bibitem[{Slack et~al.(2020)Slack, Hilgard, Singh, and
  Lakkaraju}]{Slack2020ReliableExplainability}
Dylan Slack, Sophie Hilgard, Sameer Singh, and Himabindu Lakkaraju. 2020.
\newblock \href {https://arxiv.org/abs/2008.05030} {{Reliable Post hoc
  Explanations: Modeling Uncertainty in Explainability}}.

\bibitem[{Smilkov et~al.(2017)Smilkov, Thorat, Kim, Vi{\'{e}}gas, and
  Wattenberg}]{Smilkov2017SmoothGrad:Noise}
Daniel Smilkov, Nikhil Thorat, Been Kim, Fernanda Vi{\'{e}}gas, and Martin
  Wattenberg. 2017.
\newblock \href {https://goo.gl/EfVzEE.} {{SmoothGrad: removing noise by adding
  noise}}.

\bibitem[{Socher et~al.(2013)Socher, Perelygin, Wu, Chuang, Manning, Ng, and
  Potts}]{socher-etal-2013-recursive}
Richard Socher, Alex Perelygin, Jean Wu, Jason Chuang, Christopher~D. Manning,
  Andrew Ng, and Christopher Potts. 2013.
\newblock \href {https://aclanthology.org/D13-1170} {Recursive deep models for
  semantic compositionality over a sentiment treebank}.
\newblock In \emph{Proceedings of the 2013 Conference on Empirical Methods in
  Natural Language Processing}, pages 1631--1642, Seattle, Washington, USA.
  Association for Computational Linguistics.

\bibitem[{Springenberg et~al.(2014)Springenberg, Dosovitskiy, Brox, and
  Riedmiller}]{Springenberg2014StrivingNet}
Jost~Tobias Springenberg, Alexey Dosovitskiy, Thomas Brox, and Martin
  Riedmiller. 2014.
\newblock \href {http://arxiv.org/abs/1412.6806} {{Striving for Simplicity: The
  All Convolutional Net}}.

\bibitem[{Sundararajan et~al.(2017)Sundararajan, Taly, and
  Yan}]{Sundararajan2017AxiomaticNetworks}
Mukund Sundararajan, Ankur Taly, and Qiqi Yan. 2017.
\newblock \href {http://arxiv.org/abs/1703.01365} {{Axiomatic Attribution for
  Deep Networks}}.

\bibitem[{Touvron et~al.(2023)Touvron, Lavril, Izacard, Martinet, Lachaux,
  Lacroix, Rozi{\`{e}}re, Goyal, Hambro, Azhar, Rodriguez, Joulin, Grave, and
  Lample}]{Touvron2023LLaMA:Models}
Hugo Touvron, Thibaut Lavril, Gautier Izacard, Xavier Martinet, Marie-Anne
  Lachaux, Timothee Lacroix, Baptiste Rozi{\`{e}}re, Naman Goyal, Eric Hambro,
  Faisal Azhar, Aurelien Rodriguez, Armand Joulin, Edouard Grave, and Guillaume
  Lample. 2023.
\newblock \href {https://github.com/facebookresearch/xformers} {{LLaMA: Open
  and Efficient Foundation Language Models}}.

\bibitem[{van~der Waa et~al.(2021)van~der Waa, Nieuwburg, Cremers, and
  Neerincx}]{vanderWaa2021EvaluatingExplanations}
Jasper van~der Waa, Elisabeth Nieuwburg, Anita Cremers, and Mark Neerincx.
  2021.
\newblock \href {https://doi.org/10.1016/j.artint.2020.103404} {{Evaluating
  XAI: A comparison of rule-based and example-based explanations}}.
\newblock \emph{Artificial Intelligence}, 291.

\bibitem[{Wang et~al.(2022)Wang, Xu, Liu, Cheng, and
  Li}]{Wang2022SemAttack:Spaces}
Boxin Wang, Chejian Xu, Xiangyu Liu, Yu~Cheng, and Bo~Li. 2022.
\newblock \href {http://arxiv.org/abs/2205.01287} {{SemAttack: Natural Textual
  Attacks via Different Semantic Spaces}}.

\bibitem[{Watson et~al.(2023)Watson, O'Hara, Tax, Mudd, and
  Guy}]{Watson2023ExplainingValues}
David~S Watson, Joshua O'Hara, Niek Tax, Richard Mudd, and Ido Guy. 2023.
\newblock {Explaining Predictive Uncertainty with Information Theoretic Shapley
  Values}.
\newblock \emph{37th Conference on Neural Information Processing Systems
  (NeurIPS 2023)}.

\bibitem[{Weidinger et~al.(2021)Weidinger, Mellor, Rauh, Griffin, Uesato,
  Huang, Cheng, Glaese, Balle, Kasirzadeh, Kenton, Brown, Hawkins, Stepleton,
  Biles, Birhane, Haas, Rimell, Hendricks, Isaac, Legassick, Irving, Gabriel,
  and Com>}]{Weidinger2021EthicalModels}
Laura Weidinger, John Mellor, Maribeth Rauh, Conor Griffin, Jonathan Uesato,
  Po-Sen Huang, Myra Cheng, Mia Glaese, Borja Balle, Atoosa Kasirzadeh, Zac
  Kenton, Sasha Brown, Will Hawkins, Tom Stepleton, Courtney Biles, Abeba
  Birhane, Julia Haas, Laura Rimell, Lisa~Anne Hendricks, William Isaac, Sean
  Legassick, Geoffrey Irving, Iason Gabriel, and <lweidinger@deepmind Com>.
  2021.
\newblock {Ethical and social risks of harm from Language Models}.

\bibitem[{Wiegreffe and Marasovi{\'{c}}(2021)}]{Wiegreffe2021TeachProcessing}
Sarah Wiegreffe and Ana Marasovi{\'{c}}. 2021.
\newblock \href {http://arxiv.org/abs/2102.12060} {{Teach Me to Explain: A
  Review of Datasets for Explainable Natural Language Processing}}.

\bibitem[{Yu et~al.(2024)Yu, Xue, Zhang, Wang, Liu, and
  Zhu}]{Yu2024ExploringSystems}
Xiaowei Yu, Yao Xue, Lu~Zhang, Li~Wang, Tianming Liu, and Dajiang Zhu. 2024.
\newblock {Exploring the Impact of Information Entropy Change in Learning
  Systems}.
\newblock \emph{Accepted to ICLR 2024}.

\bibitem[{Zhang et~al.(2022{\natexlab{a}})Zhang, Roller, Goyal, Artetxe, Chen,
  Chen, Dewan, Diab, Li, Lin, Mihaylov, Ott, Shleifer, Shuster, Simig, Koura,
  Sridhar, Wang, Zettlemoyer, and Ai}]{ZhangOPT:Models}
Susan Zhang, Stephen Roller, Naman Goyal, Mikel Artetxe, Moya Chen, Shuohui
  Chen, Christopher Dewan, Mona Diab, Xian Li, Victoria Lin, Todor Mihaylov,
  Myle Ott, Sam Shleifer, Kurt Shuster, Daniel Simig, Singh Koura, Anjali
  Sridhar, Tianlu Wang, Luke Zettlemoyer, and Meta Ai. 2022{\natexlab{a}}.
\newblock \href {https://bigscience.} {{OPT: Open Pre-trained Transformer
  Language Models}}.

\bibitem[{Zhang et~al.(2019)Zhang, Song, Sun, Tan, and
  Udell}]{Zhang2019WhyExplanations}
Yujia Zhang, Kuangyan Song, Yiming Sun, Sarah Tan, and Madeleine Udell. 2019.
\newblock {"Why Should You Trust My Explanation?" Understanding Uncertainty in
  LIME Explanations}.
\newblock \emph{International Conference on Machine Learning AI for Social Good
  Workshop.}

\bibitem[{Zhang et~al.(2022{\natexlab{b}})Zhang, Pan, Tan, and
  Kan}]{Zhang2022InterpretingPerturbations}
Yunxiang Zhang, Liangming Pan, Samson Tan, and Min-Yen Kan. 2022{\natexlab{b}}.
\newblock {Interpreting the Robustness of Neural NLP Models to Textual
  Perturbations}.

\end{thebibliography}
\bibliographystyle{acl_natbib}

\appendix

\section{Model training specifications}
\label{sec:APP_searchspace}
The pre-trained models are connected to a classification head and fine-tuned on the datasets listed in Table \ref{tab:Datasets} using either previously reported optimal hyperparameters or with hyperparameters we identified by exploring the search space with raytuning \cite{liaw2018tune}. We use pre-trained tokenizers specific to each model. For BERT, we rely on BERT\textsubscript{base},, which is 110 million parameters. We use RoBERTa\textsubscript{base}, which is 125 million parameters. ELECTRA is 110 million parameters. We rely on GPT2\textsubscript{medium}, which is 345 million parameters, and OPT-350M, which is 350 million parameters. BERT, RoBERTa, and ELECTRA are trained and assessed on Titan RTX GPUs; GPT2 and OPT are trained and assessed on A100 GPUs.

\subsection{SST-2}

Our BERT model uses the hyperparameters reported by the best-performing BERT-base model on the SST-2 task, which achieves 92.3\% accuracy on the evaluation set\footnote{https://huggingface.co/gchhablani/bert-base-cased-finetuned-sst2}.
While we cannot find hyperparameters reaching the performance described in the original RoBERTa-base (94.8\%) article \cite{Liu2019RoBERTa:Approach}, we choose the hyperparameters specified by this model card \footnote{https://huggingface.co/Bhumika/RoBERTa-base-finetuned-sst2}, which achieves an accuracy of 94.5\% on the evaluation set.
Our ELECTRA model uses the best-performing hyperparameters listed in the original article \cite{Clark2020ELECTRA:GENERATORS}, which achieves an accuracy of 96.0\% on the evaluation set.
Our GPT2 model uses the hyperparameters listed in the original article \cite{Radford2019LanguageLearners} and achieves an accuracy of 92\% on the evaluation set. For OPT, we used the hyperparameters specified by the huggingface model card\footnote{https://huggingface.co/tianyisun/opt-350m-finetuned-sst2}, which achieved an accuracy of 91\% on the evaluation set.

\subsection{SemEval, HateXplain} %
Model hyperparameters are identified using a hyperparameter search space with a learning rate between $1e-6$ and $1e-4$, epochs between 1 and 10, and a batch size of (4, 8, 16, 32).

Our final hyperparameters for SemEval, and HateXplain are shown in Tables \ref{tab:hyper_Semeval} and \ref{tab:hyper_hx}.

\begin{table}[ht]
\small
\begin{adjustbox}{max width=1.0\columnwidth,center}
\begin{tblr}{
  row{1} = {c},
  cell{1}{1} = {c=6}{},
  vline{2} = {2-16}{},
  hline{1,17} = {-}{0.08em},
  hline{2} = {-}{0.05em},
  hline{3} = {1-5}{0.03em},
  hline{3} = {6}{},
columns = {colsep=2pt},
}
\textbf{SemEval}     &               &                  &                  &               &              \\
\textbf{model}       & \textbf{BERT} & \textbf{RoBERTa} & \textbf{ELECTRA} & \textbf{GPT2} & \textbf{OPT} \\
Learning rate        & 1e-5          & 1e-5             & 3e-6             & 8e-5          & 7e-6         \\
Batch size           & 16            & 16               & 8                & 32            & 32           \\
Epochs               & 3             & 3                & 5                & 7             & 1            \\
Random seed          & 37            & 37               & 24               & 42            & 42           \\
Adam $\epsilon$      & 1e-8          & 1e-8             & 1e-8             & 1e-8          & 1e-8         \\
Adam $\beta$ 1       & 0.9           & 0.9              & 0.9              & 0.9           & 0.9          \\
Adam $\beta$ 2       & 0.999         & 0.999            & 0.999            & 0.999         & 0.999        \\
LLRD                 & None          & None             & None             & None          & None         \\
Decay type           & Linear        & Linear           & Linear           & Cosine        & Cosine       \\
{Warmup\\Fraction}   & 0             & 0                & 0                & 0.01          & 0.01         \\
{Attention\\Dropout} & 0.1           & 0.1              & 0.1              & 0.1           & 0.1          \\
Dropout              & 0.1           & 0.1              & 0.1              & 0.1           & 0.1          \\
Weight Decay         & 0             & 0                & 0                & 0.1           & 0.1          \\
Test Accuracy        & 92\%          & 94\%             & 91\%             & 91\%          & 91\%         
\end{tblr}
\end{adjustbox}
\caption{Final hyperparameters for all investigated models on the SemEval dataset}
\label{tab:hyper_Semeval}
\end{table}

\begin{table}[ht]
\small
\begin{adjustbox}{max width=1.0\columnwidth,center}
\begin{tblr}{
  row{1} = {c},
  cell{1}{1} = {c=6}{},
  vline{2} = {2-16}{},
  hline{1,17} = {-}{0.08em},
  hline{2} = {-}{0.05em},
  hline{3} = {1-5}{0.03em},
  hline{3} = {6}{},
  columns = {colsep=2pt},
}
\textbf{HateXplain}  &               &                  &                  &               &              \\
\textbf{model}       & \textbf{BERT} & \textbf{RoBERTa} & \textbf{ELECTRA} & \textbf{GPT2} & \textbf{OPT} \\
Learning rate        & 2e-5          & 6e-6             & 2e-5             & 5e-5          & 9e-6         \\
Batch size           & 32            & 32               & 8                & 32            & 8            \\
Epochs               & 5             & 5                & 2                & 6             & 1            \\
Random seed          & 2             & 2                & 6                & 42            & 42           \\
Adam $\epsilon$      & 1e-8          & 1e-8             & 1e-8             & 1e-8          & 1e-8         \\
Adam $\beta$ 1       & 0.9           & 0.9              & 0.9              & 0.9           & 0.9          \\
Adam $\beta$ 2       & 0.999         & 0.999            & 0.999            & 0.999         & 0.999        \\
LLRD                 & None          & None             & None             & None          & None         \\
Decay type           & Linear        & Linear           & Linear           & Cosine        & Cosine       \\
{Warmup\\Fraction}   & 0             & 0                & 0                & 0.01          & 0.01         \\
{Attention\\Dropout} & 0.1           & 0.1              & 0.1              & 0.1           & 0.1          \\
Dropout              & 0.1           & 0.1              & 0.1              & 0.1           & 0.1          \\
Weight Decay         & 0             & 0                & 0                & 0.1           & 0.1          \\
Test Accuracy        & 68\%          & 69\%             & 70\%             & 66\%          & 69\%         
\end{tblr}
\end{adjustbox}

\caption{Final hyperparameters for all investigated models on the HateXplain dataset}
\label{tab:hyper_hx}
\end{table}

\section{Perturbation specifications}
\subsection{Proportion replaced}
\label{sec:APP_prop}
When perturbing a text by $\alpha$\, $\alpha$ indicates the proportion of the text that is modified by a perturbation type. Texts are never fully perturbed, so, if $\alpha=0.95$, and the length of the text ($N_{tokens}$) is fewer than 20 tokens, at least one token is left unmodified. For very short texts, as seen in SST-2, the data point is left unmodified until $\alpha \times N_{tokens} \geq 1$. Effects of perturbation level are only assessed at an aggregated level to visualize the rate of metric change with increasing perturbation.

\subsection{Synonym replacement}
\label{sec:APP_synonyms}
Across all synonym replacements, we preserve the case of the original word (e.g. HAPPY! becomes GLAD!). In addition, we use NLTK POS tagger to tag each word to a part of speech for more precise synonym mapping. If NLTK is unable to find a part of speech, or it must be dropped when merging multiple tokens (e.g. if one token is not a punctuation mark or a possession-indicator), then we ignore part of speech.

We followed the following hierarchical rules for synonym replacement:

1. Tokens beginning with \verb|http://t.co/| or \verb|https://t.co/| are replaced with a similar randomly-generated URL string following a similar regex pattern

2. Tokens beginning with a \#, we remove the \#, find a synonym, and then re-add the \#.

3. Tokens beginning with a @ are replaced with another random Twitter ID found in the test set.

4. Determinants are replaced another random determinant (\verb|['a', 'an', 'the', 'this', 'that']|). Similarly question determinants are replaced with other question determinants.  (\verb|['that', 'what', 'whatever', 'which',|
\verb|'whichever']|)

5. Proper nouns are replaced with a randomly generated first name or last name. If the original name ends with a "'s", this is removed and then re-added to the synonym.

6. If the word is a quote \verb|[ "'", "''", "`", "``", '"']|, bracket \verb|["(", ")", "{", "}", "[", "]", '/']|, punctuation mark \verb|[ '.', '!', '?', ',']|, or sentence break \verb|['-', '--', ',', ':', ';']|, it is replaced by another quote, bracket, punctuation mark or sentence break.

7. If the word is an arabic number (e.g. 7), it is replaced by its english equivalent (e.g. seven).

8. If a word has a synonym in WordNet or a word with an \verb|Equivalence| relation in PPDB 2.0, we randomly select a synonym from the set. If a synonym is longer than one word, the words are hyphenated (This is done to simplify matching of saliency maps between perturbations).

9. If the word starts or ends with a quote, bracket, punctuation mark or line break, we remove the character, find a synonym and then re-add the character in question.

10. If there are hyphens, periods or '//' spaced throughout the word, we use the punctuation mark to parse the word and find a replacement word for one of the word subsections.

11. If a word has a forward or reverse entailment in PPDB 2.0, we randomly choose one as a replacement. (e.g. berry for fruit or fruit for berry).

12. If no synonym has been found with using POS tags, I will expand my search in WordNet and PPDB 2.0 without the POS tag. 

13. If the word ends with the popular suffixes '-ish', '-ness', or '-less', we remove the suffix, find a synonym, and then re-add the suffix in question.

\section{Model and dataset-level differences}
While the results for BERT and SST-2 are visualized in the article, we provide the results for all investigated models and datasets below.

\subsection{The effect of perturbation hierarchy on uncertainty and explanations}
\label{app:RQ1-1_all}

\begin{figure*}
    \centering
    \begin{adjustbox}{max width=1.0\textwidth,center}
    \includegraphics[scale=0.35]{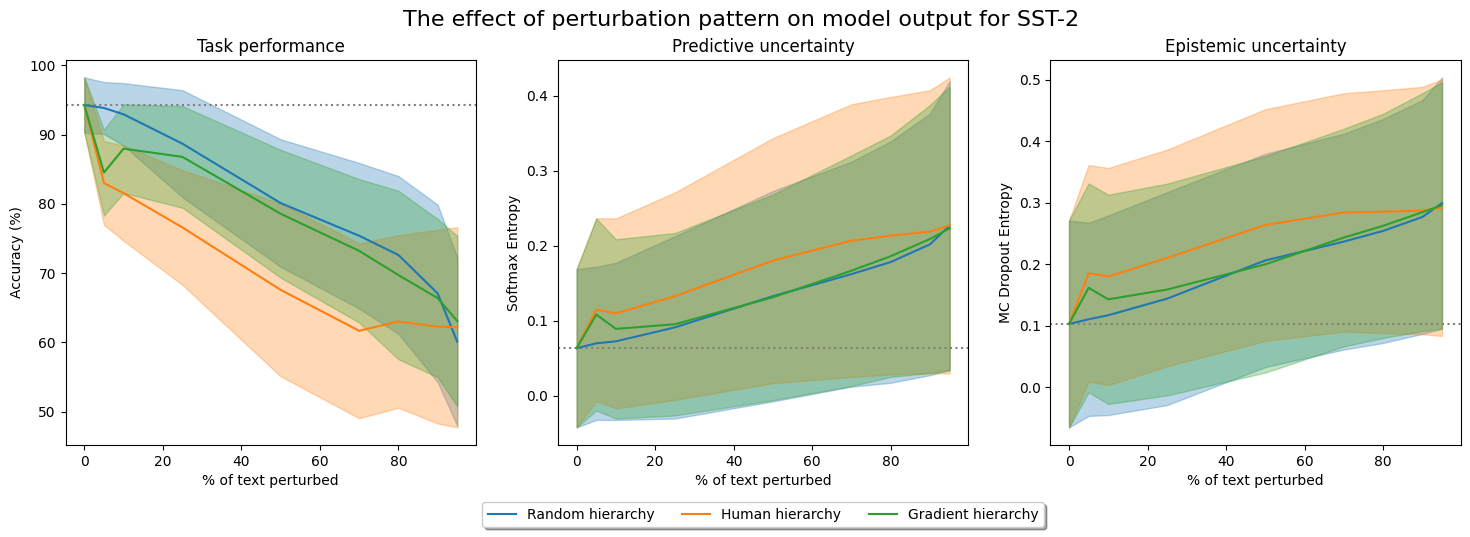}\end{adjustbox}\\
    \begin{adjustbox}{max width=1.0\textwidth,center}
    \includegraphics[scale=0.35]{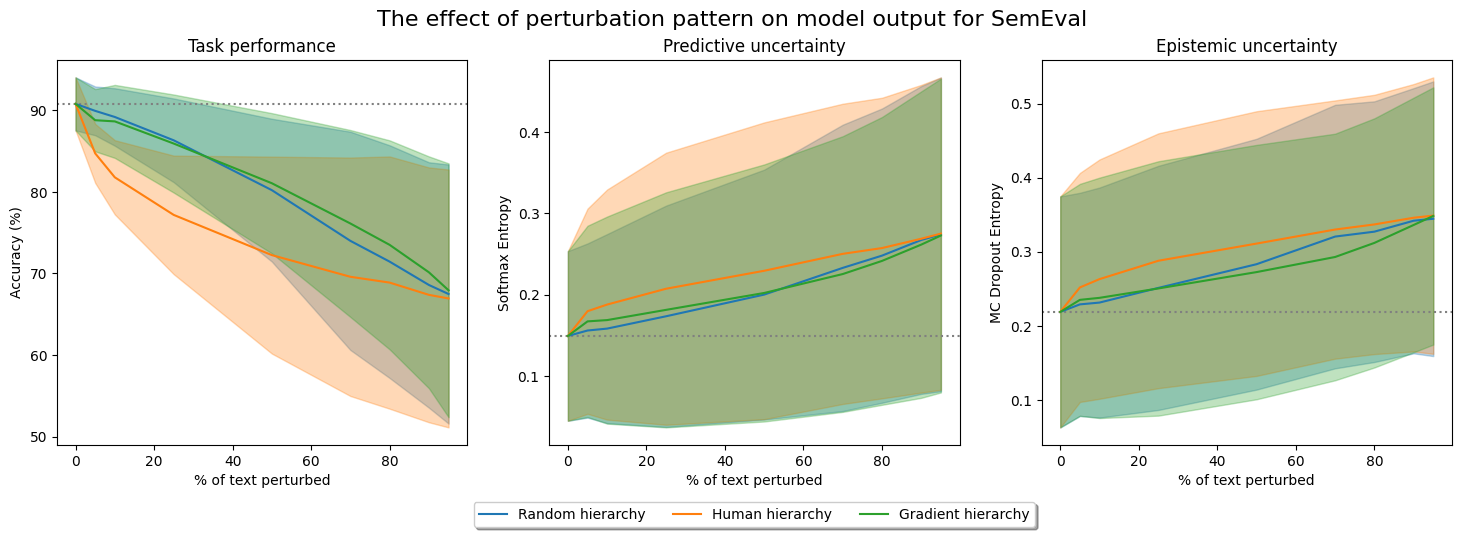}\end{adjustbox}\\
    \begin{adjustbox}{max width=1.0\textwidth,center}
    \includegraphics[scale=0.35]{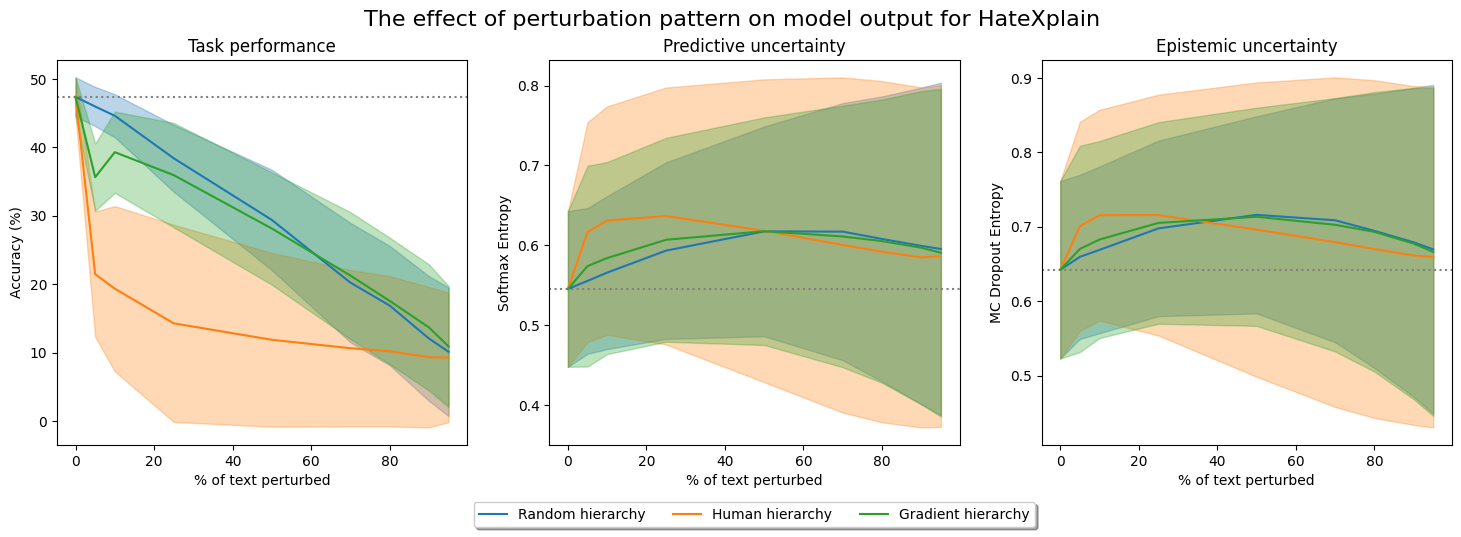}\end{adjustbox}\\
    \caption{We show the differential effect of our perturbation hierarchies across the different datasets investigated. Values are averaged over all 7 perturbation types.}
    \label{fig:RQ1-1_all}
\end{figure*}

We show the results of our investigations into the effect of perturbation hierarchies (see \S\ref{sec:RQ1_res}) across our investigated datasets in Figure \ref{fig:RQ1-1_all}. \textbf{Results:} Across all 4 datasets, we find that human-hierarchied perturbation has the strongest impact on task performance, uncertainty, and explanation plausibility. Furthermore, we can see that, while gradient and random-hierarchical perturbation has typically quite similar impact, the difference is greatest at low ($\alpha=.05$) levels of perturbation, and begins to diminish at higher levels ($\alpha=.1$). Interestingly, we see high levels of perturbation have a parabolic relationship with uncertainty in the HateXplain dataset.

\subsection{The effect of perturbation type on model output}
\label{app:RQ1-2_all}

\begin{figure*}
    \centering
    \begin{adjustbox}{max width=1.0\textwidth,center}
    \includegraphics[scale=0.35]{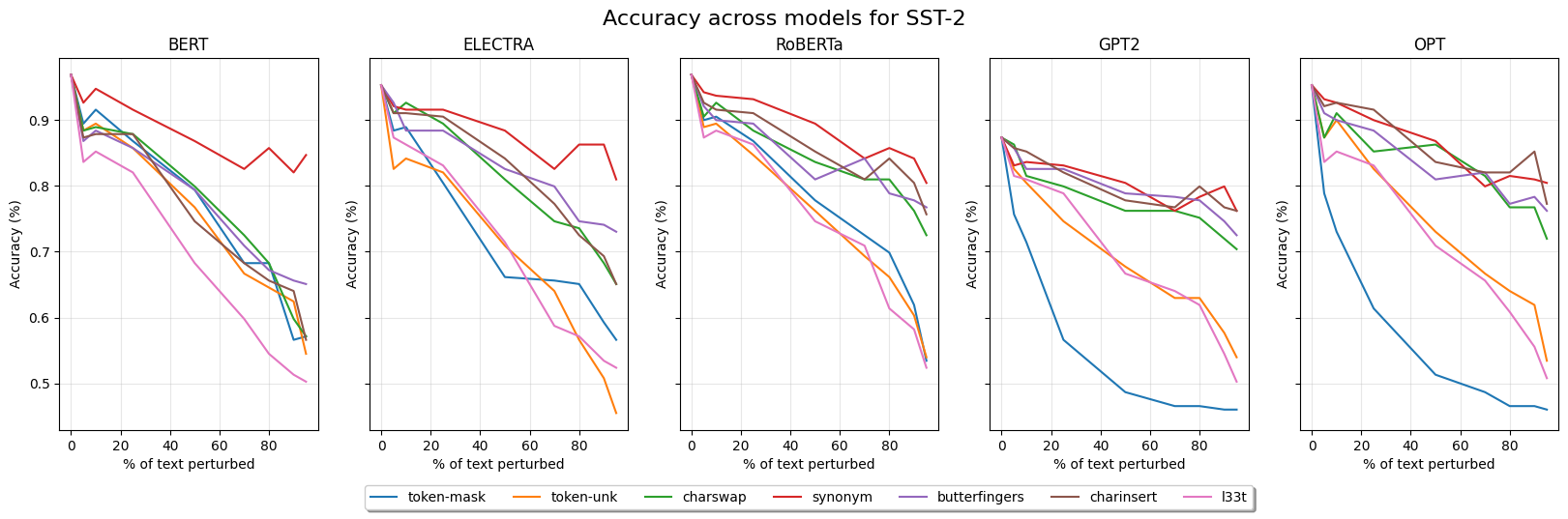}\end{adjustbox}\\
    \begin{adjustbox}{max width=1.0\textwidth,center}
    \includegraphics[scale=0.35]{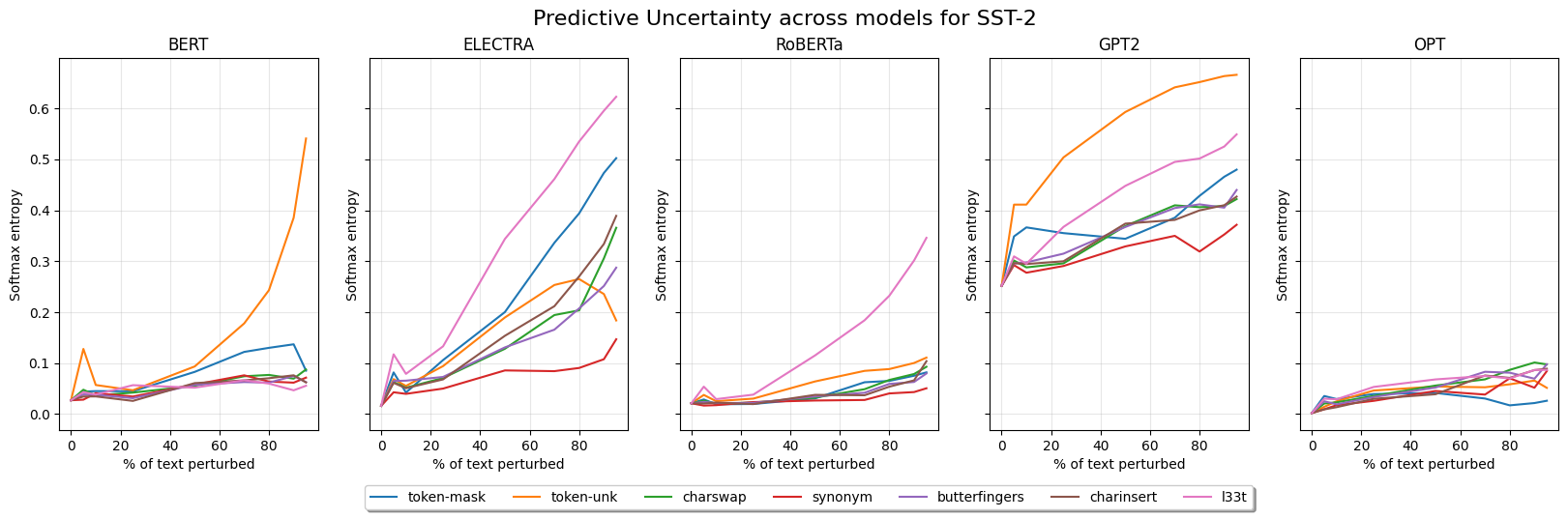}\end{adjustbox}\\
    \begin{adjustbox}{max width=1.0\textwidth,center}
    \includegraphics[scale=0.35]{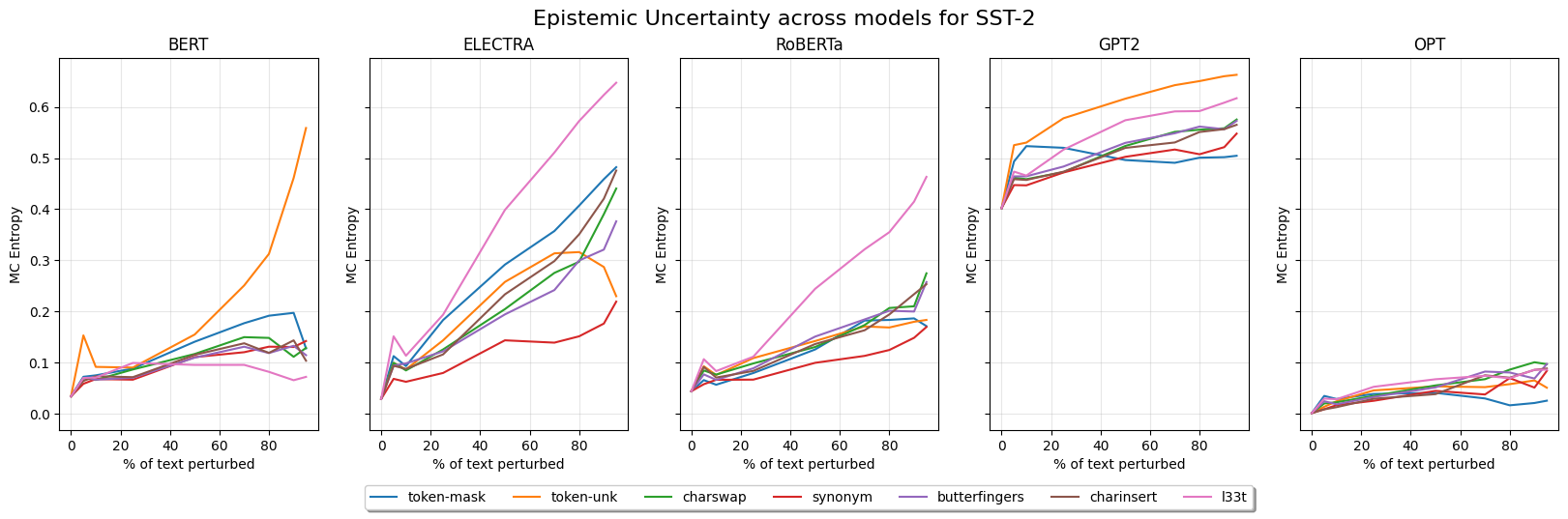}\end{adjustbox}\\
    
    \caption{We show the differential effect of increasing levels of text perturbation on model accuracy and both measures of uncertainty on the \textbf{SST-2} dataset. Values are averaged over all hierarchies.}
    \label{fig:TASK_DIFF_SST}
\end{figure*}

\begin{figure*}
    \centering
    \begin{adjustbox}{max width=1.0\textwidth,center}
    \includegraphics[scale=0.35]{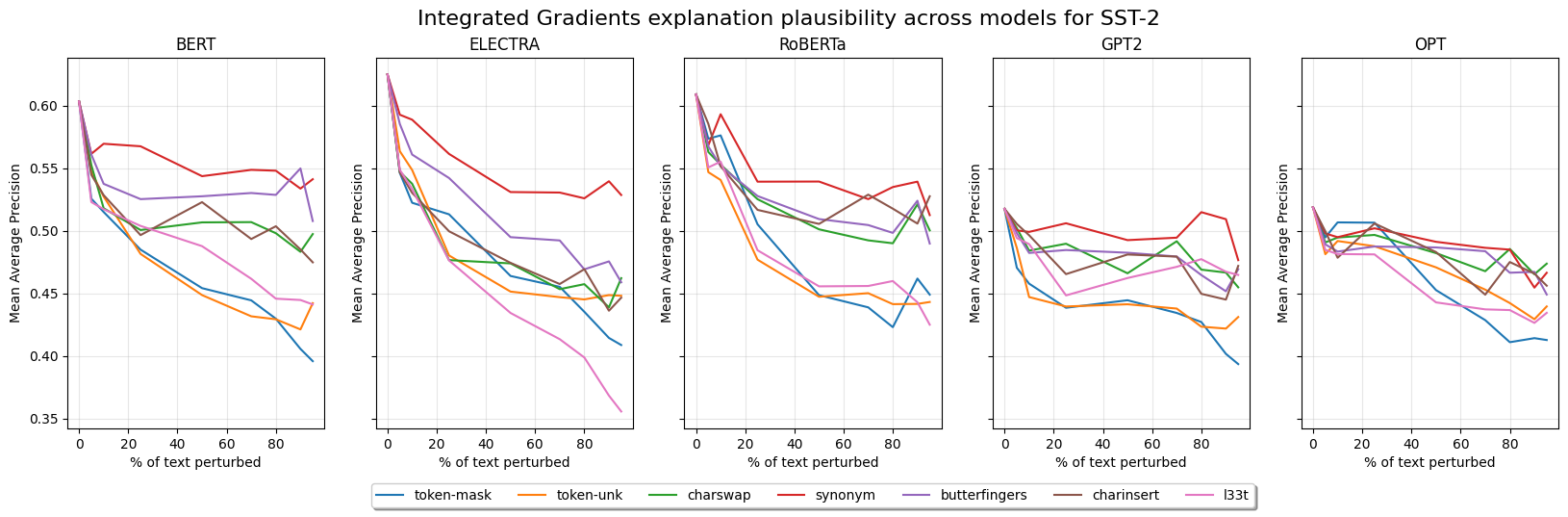}\end{adjustbox}\\
    \begin{adjustbox}{max width=1.0\textwidth,center}
    \includegraphics[scale=0.35]{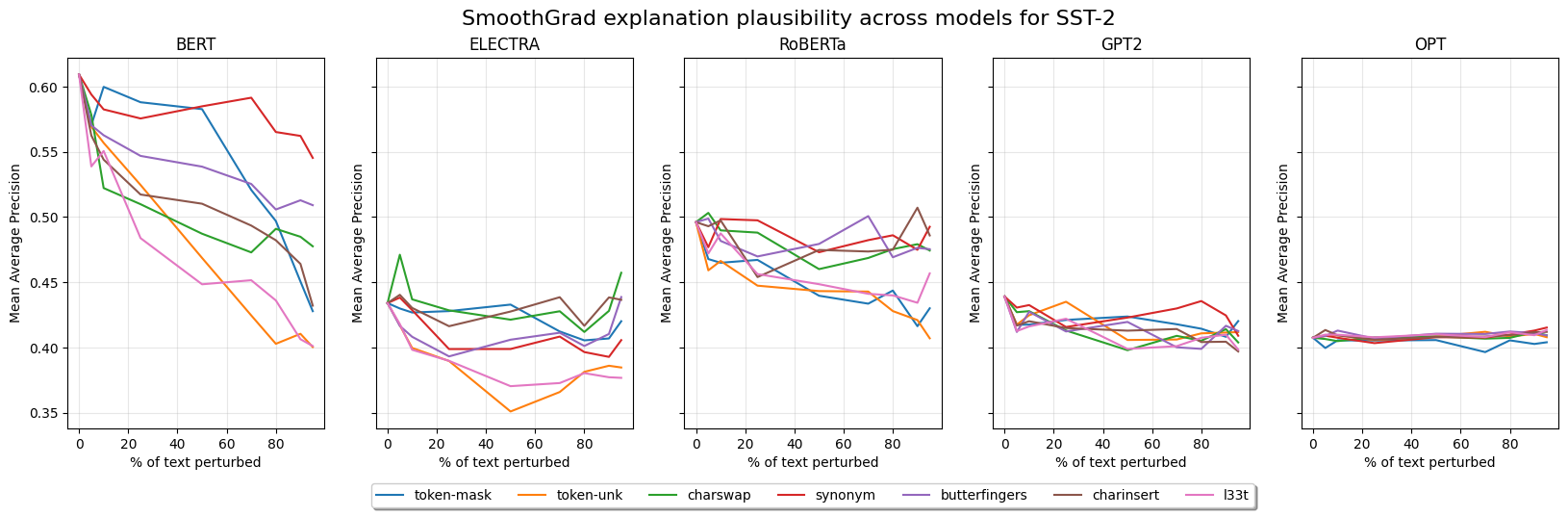}\end{adjustbox}\\
    \begin{adjustbox}{max width=1.0\textwidth,center}
    \includegraphics[scale=0.35]{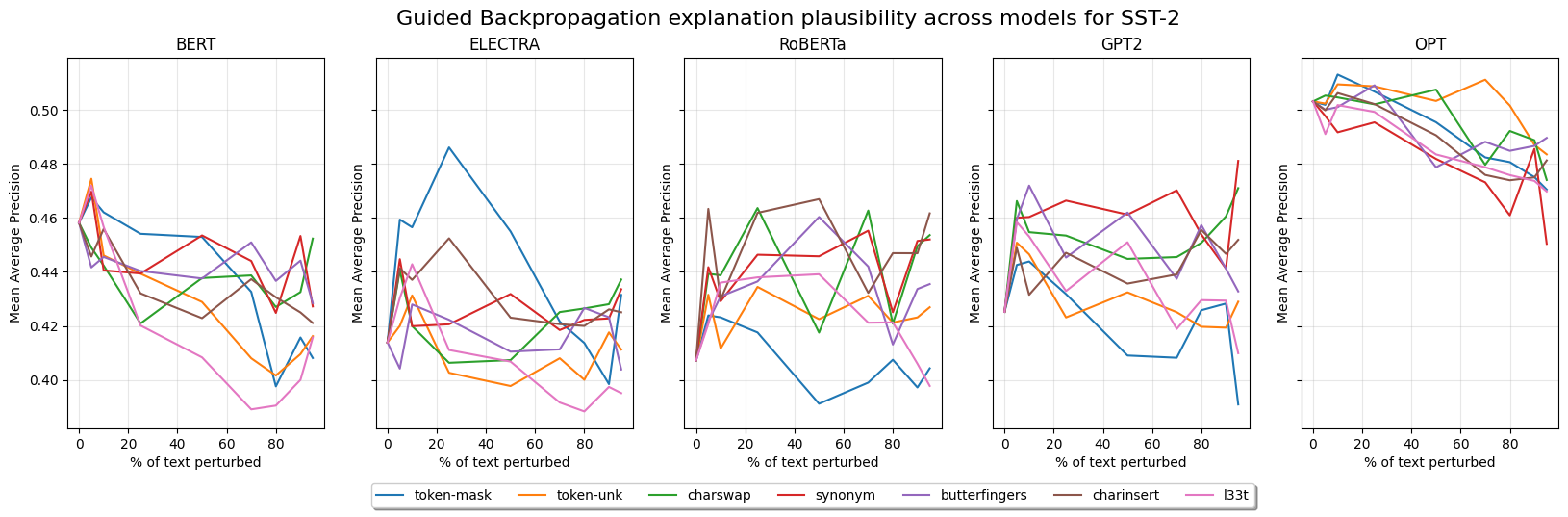}\end{adjustbox}\\
    \begin{adjustbox}{max width=1.0\textwidth,center}
    \includegraphics[scale=0.35]{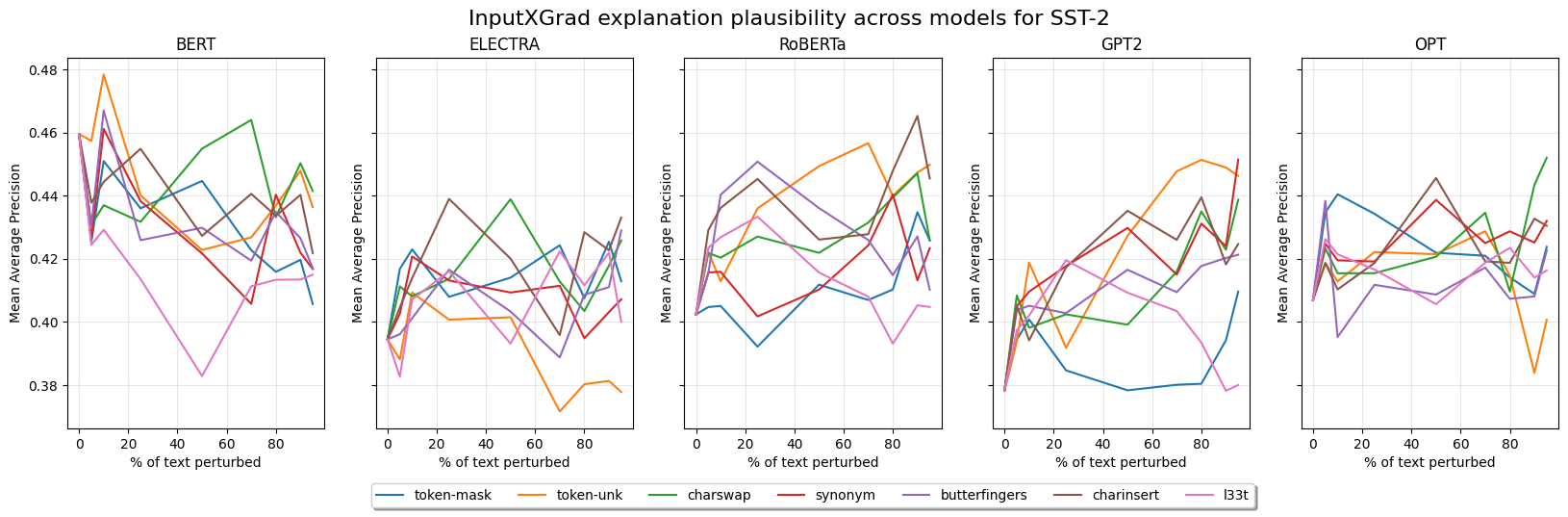}\end{adjustbox}\\
    
    \caption{We show the differential effect of increasing levels of text perturbation on the explanation plausiblity of all saliency maps \textbf{SST-2} dataset. Values are averaged over all hierarchies.}
    \label{fig:TASK_DIFF_SST_COH}
\end{figure*}

\begin{figure*}
    \centering
    \begin{adjustbox}{max width=1.0\textwidth,center}
    \includegraphics[scale=0.35]{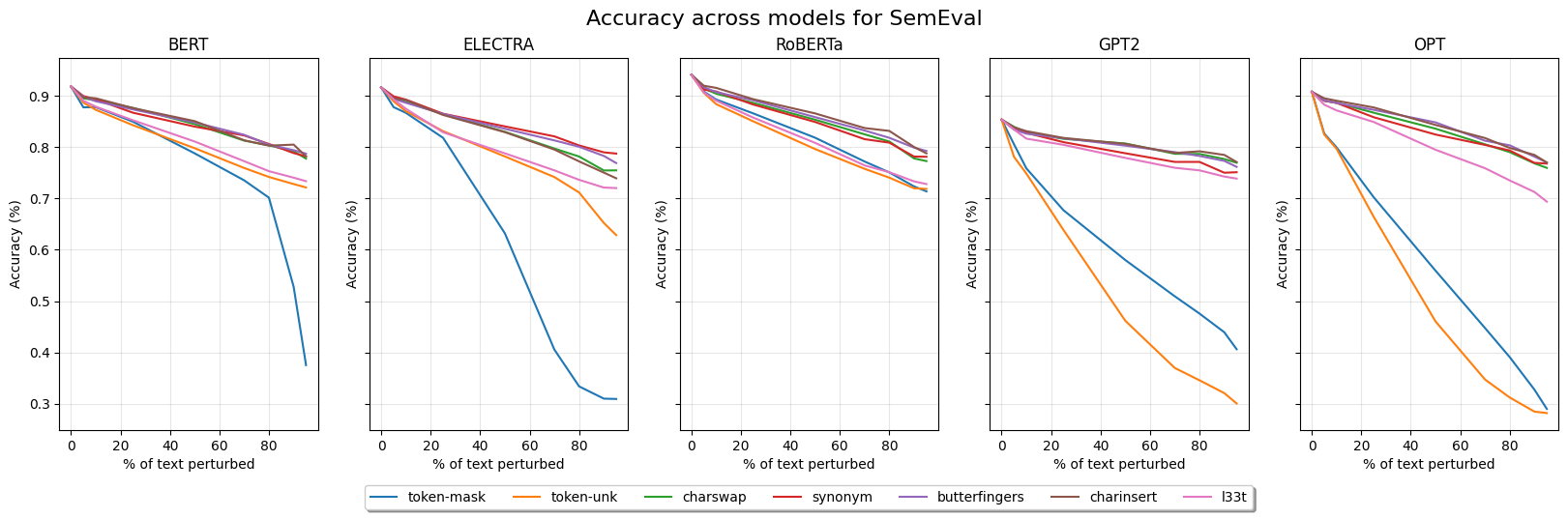}\end{adjustbox}\\
    \begin{adjustbox}{max width=1.0\textwidth,center}
    \includegraphics[scale=0.35]{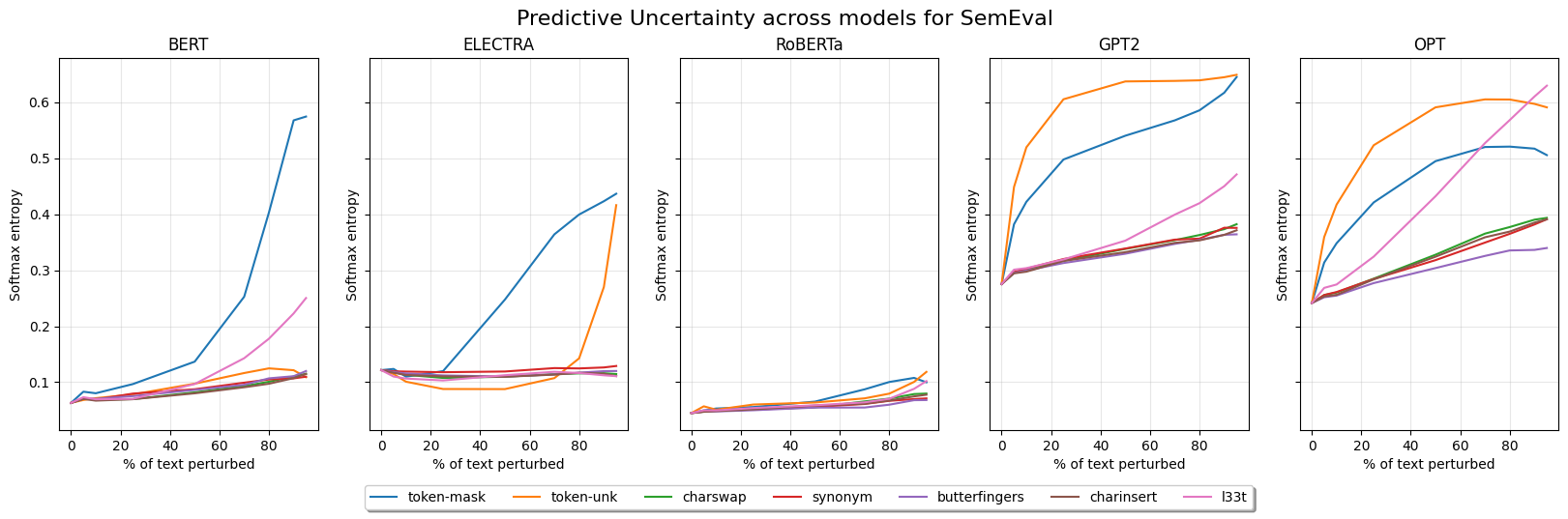}\end{adjustbox}\\
    \begin{adjustbox}{max width=1.0\textwidth,center}
    \includegraphics[scale=0.35]{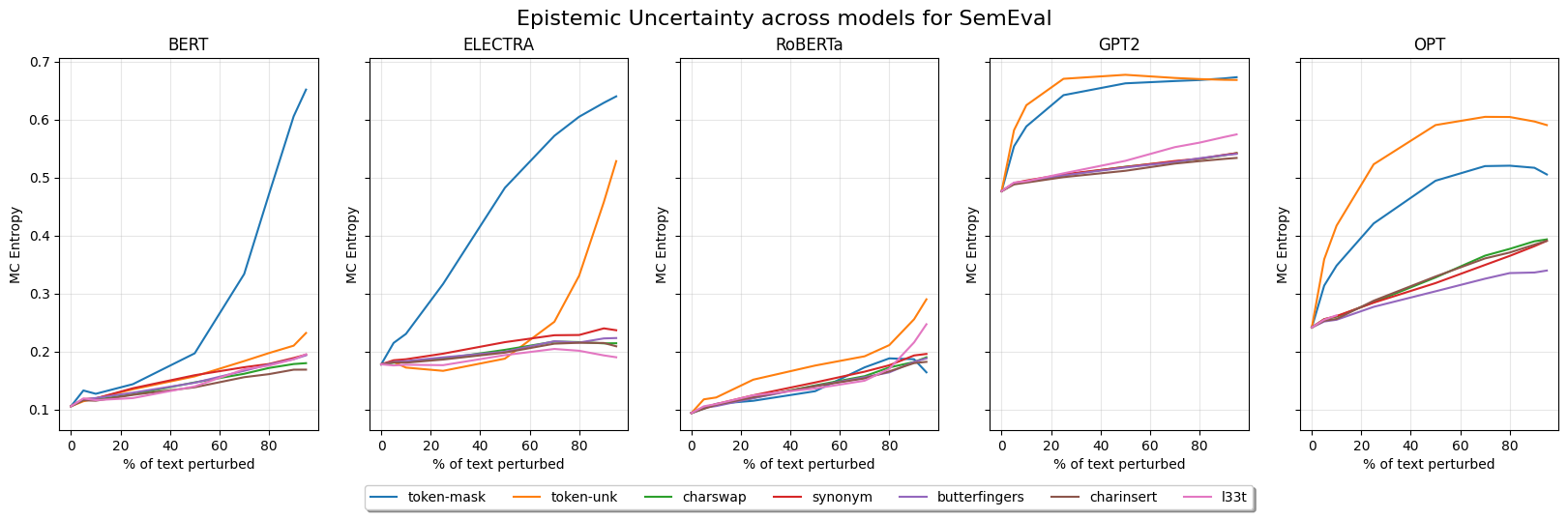}\end{adjustbox}\\
    
    \caption{We show the differential effect of increasing levels of text perturbation on model accuracy and both measures of uncertainty on the \textbf{SemEval} dataset. Values are averaged over all hierarchies.}
    \label{fig:TASK_DIFF_SEM}
\end{figure*}

\begin{figure*}
    \centering
    \begin{adjustbox}{max width=1.0\textwidth,center}
    \includegraphics{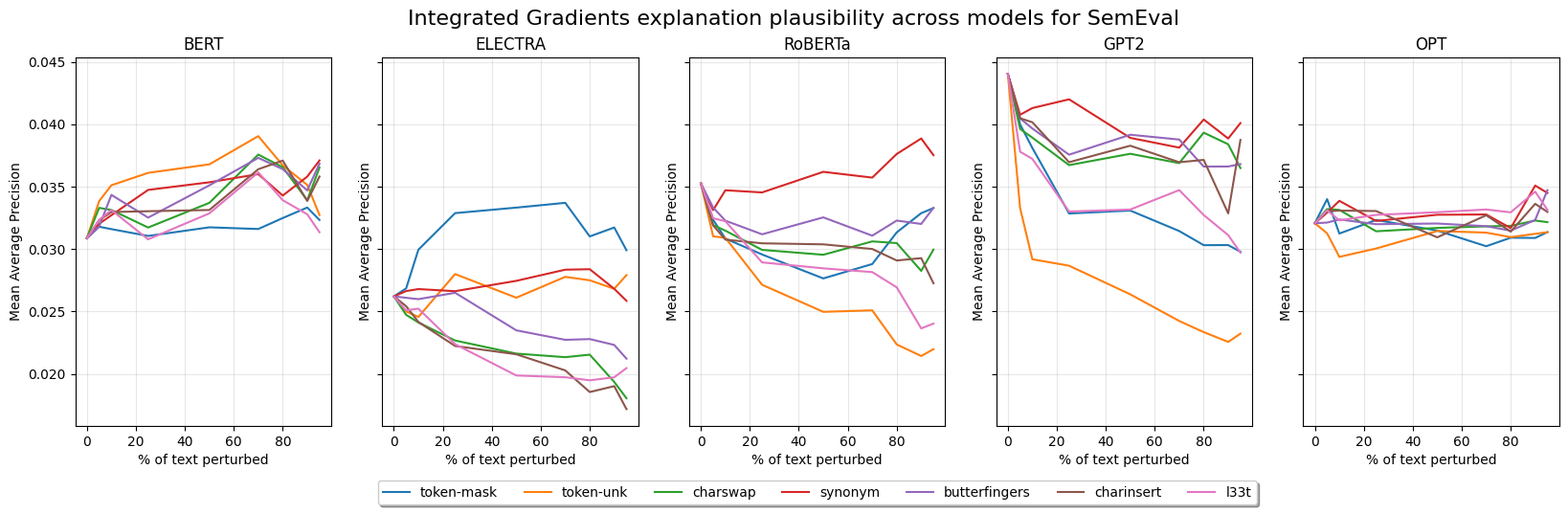}\end{adjustbox}\\
    \begin{adjustbox}{max width=1.0\textwidth,center}
    \includegraphics{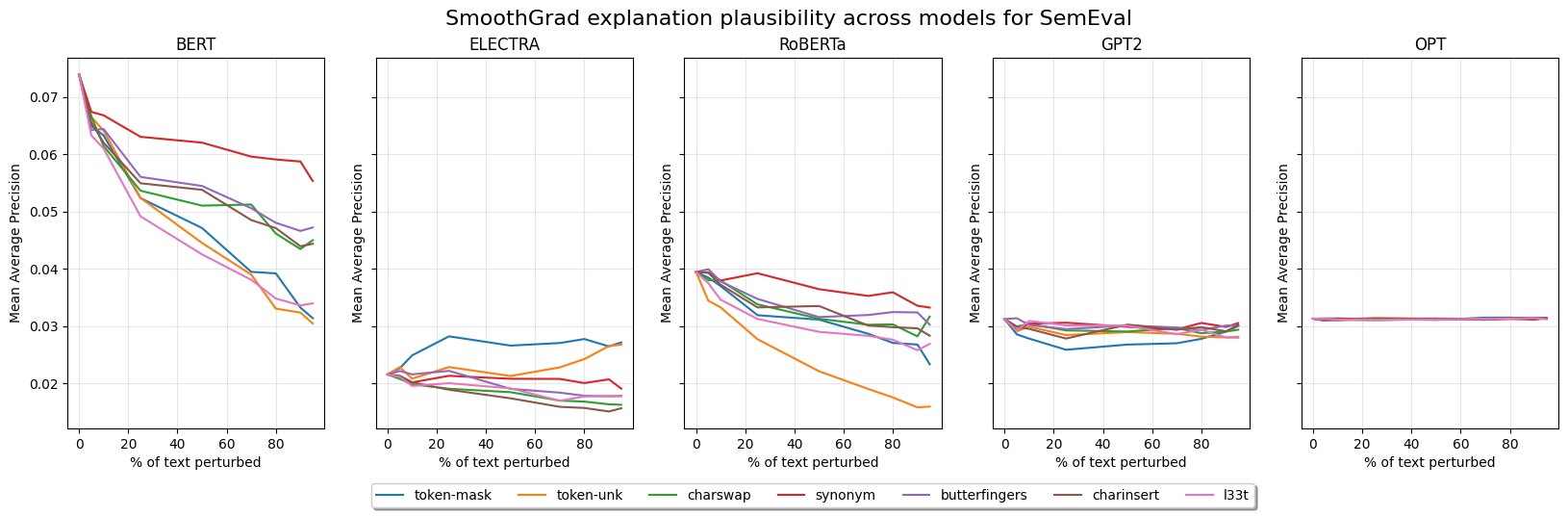}\end{adjustbox}\\
    \begin{adjustbox}{max width=1.0\textwidth,center}
    \includegraphics{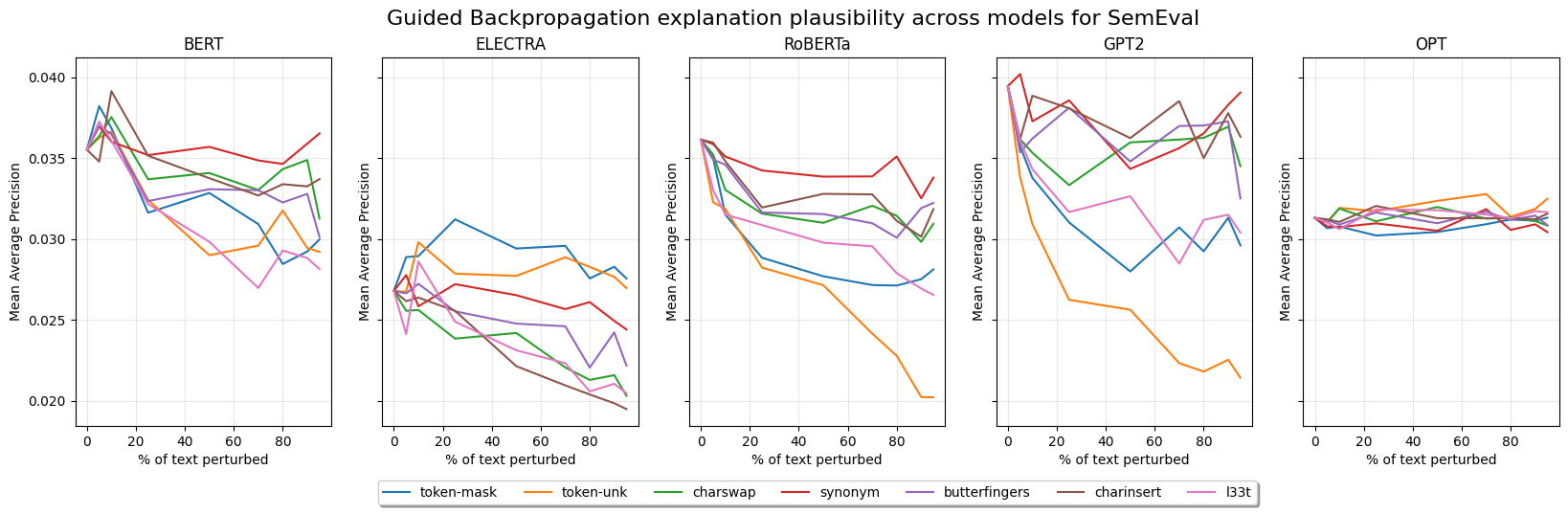}\end{adjustbox}\\
    \begin{adjustbox}{max width=1.0\textwidth,center}
    \includegraphics{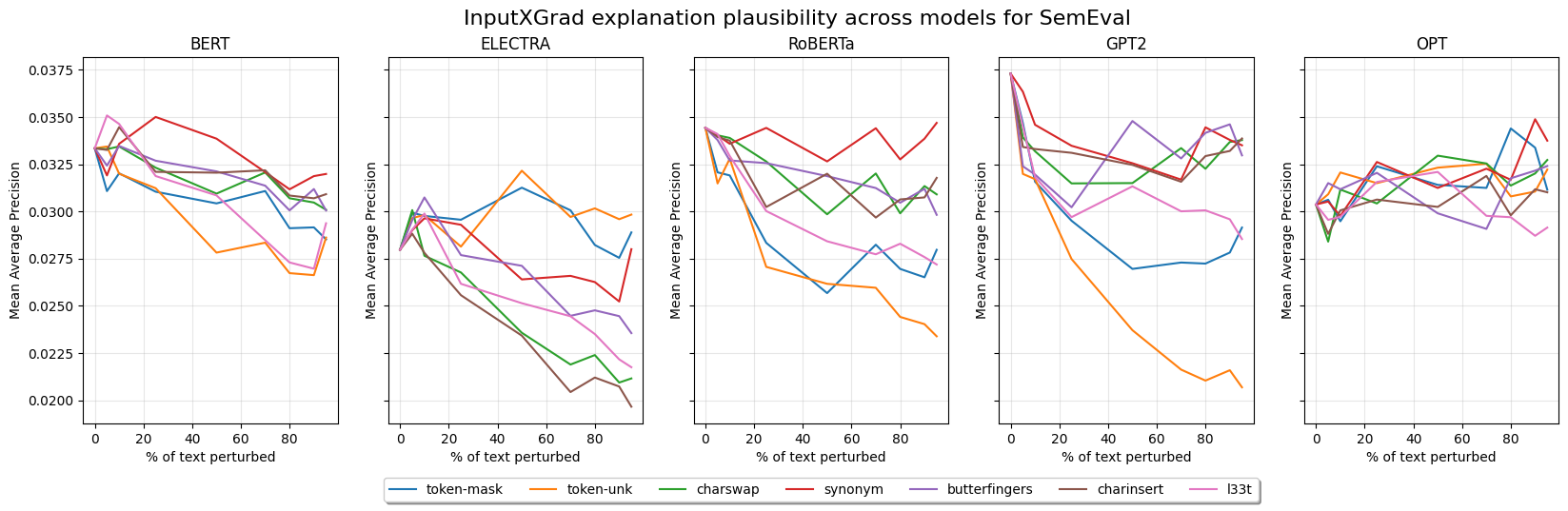}\end{adjustbox}\\
    
    \caption{We show the differential effect of increasing levels of text perturbation on the explanation plausiblity of all saliency maps \textbf{Semeval} dataset. Values are averaged over all hierarchies.}
    \label{fig:TASK_DIFF_SEM_COH}
\end{figure*}

\begin{figure*}
    \centering
    \begin{adjustbox}{max width=1.0\textwidth,center}
    \includegraphics[scale=0.35]{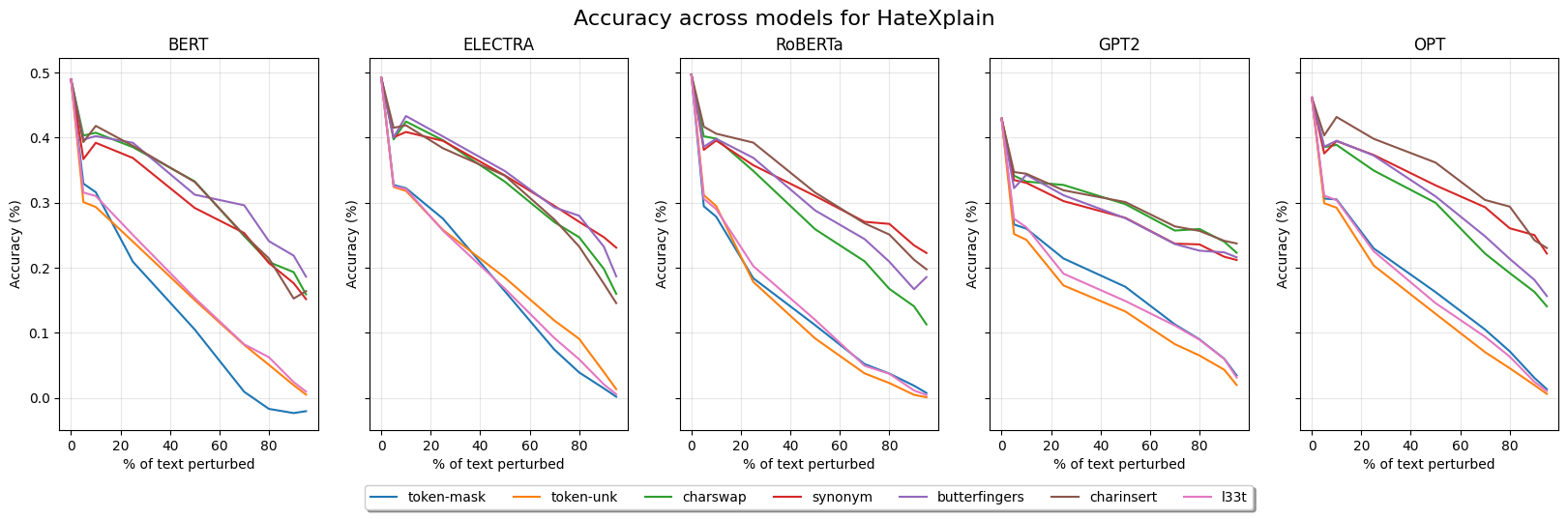}\end{adjustbox}\\
    \begin{adjustbox}{max width=1.0\textwidth,center}
    \includegraphics[scale=0.35]{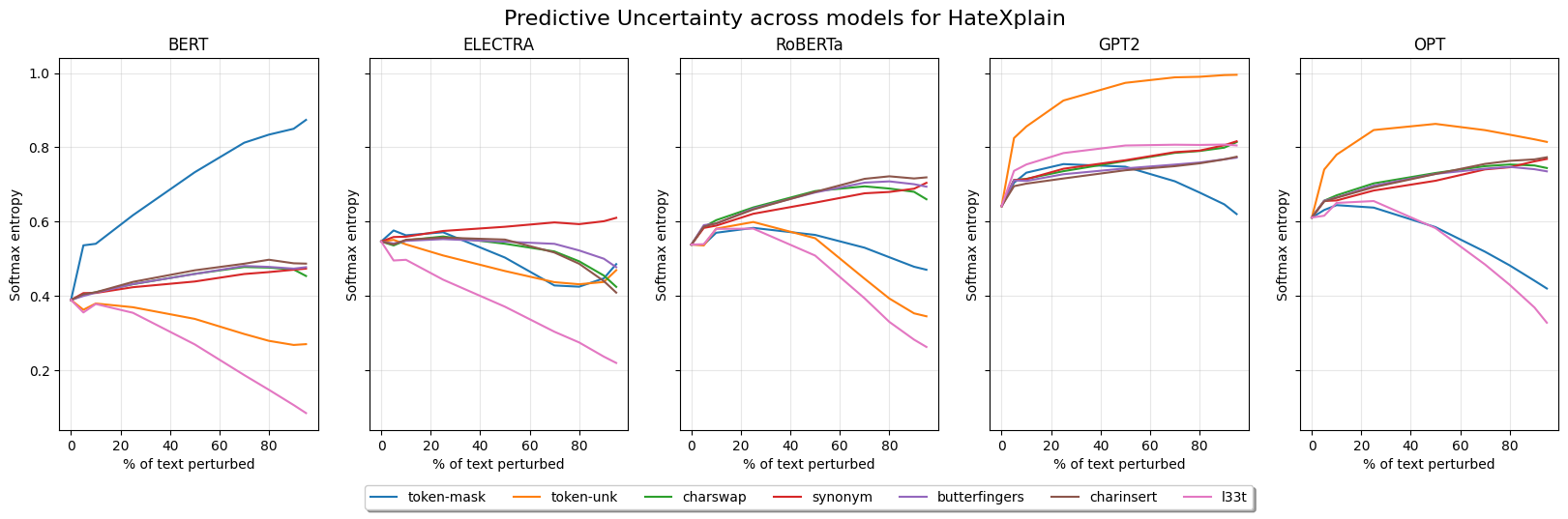}\end{adjustbox}\\
    \begin{adjustbox}{max width=1.0\textwidth,center}
    \includegraphics[scale=0.35]{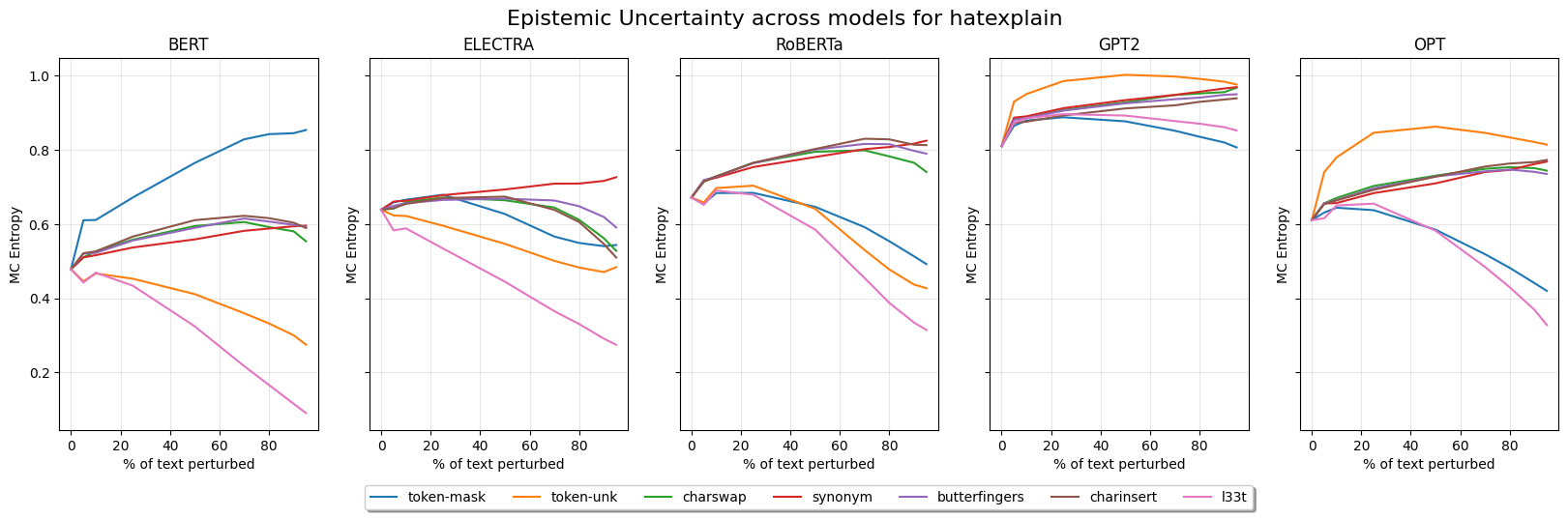}\end{adjustbox}\\
    
    \caption{We show the differential effect of increasing levels of text perturbation on model accuracy and both measures of uncertainty on the \textbf{HateXplain} dataset. Values are averaged over all hierarchies.}
    \label{fig:TASK_DIFF_HX}
\end{figure*}

\begin{figure*}
    \centering
    \begin{adjustbox}{max width=1.0\textwidth,center}
    \includegraphics[scale=0.35]{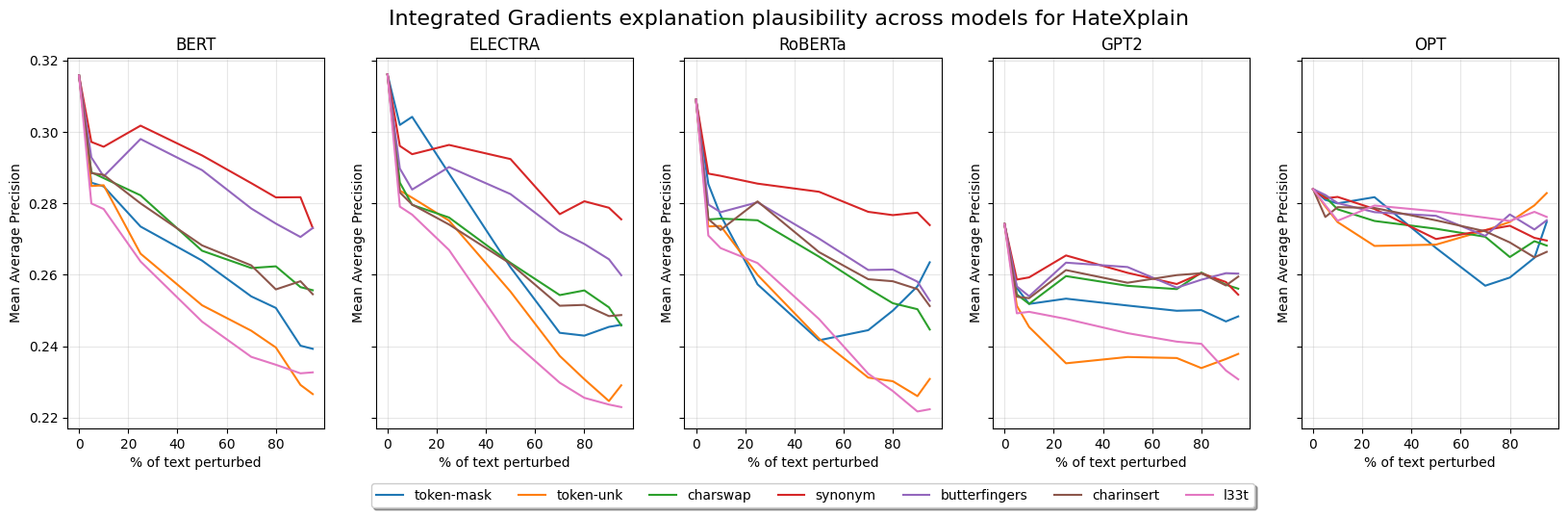}\end{adjustbox}\\
    \begin{adjustbox}{max width=1.0\textwidth,center}
    \includegraphics[scale=0.35]{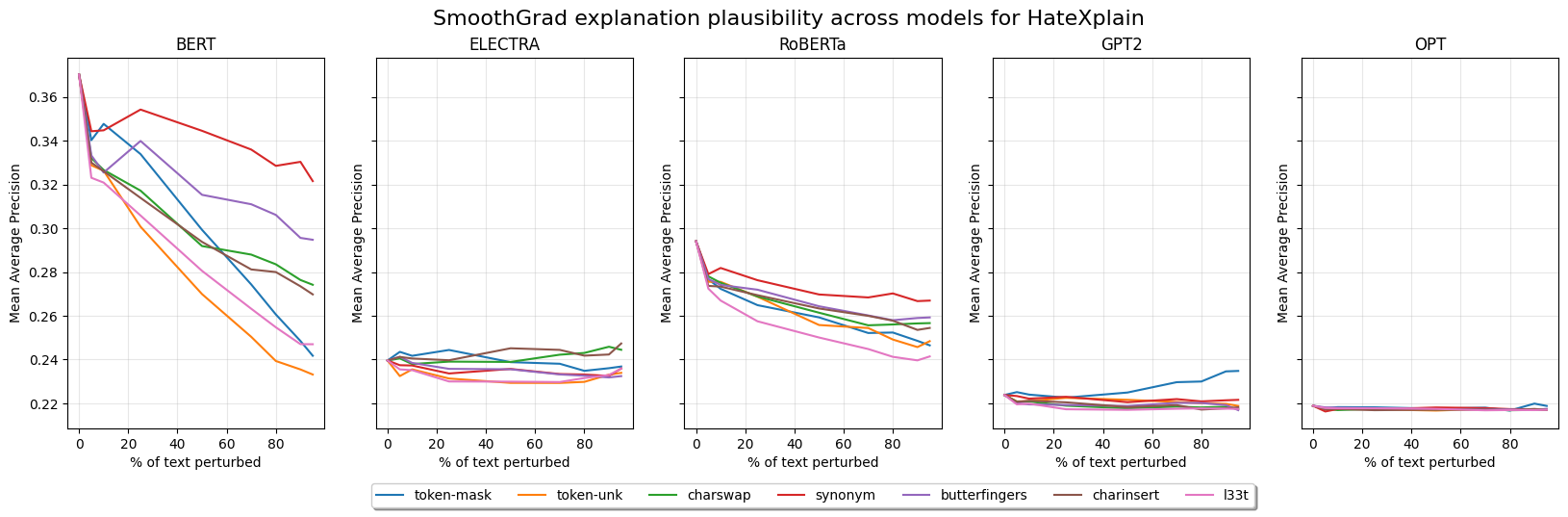}\end{adjustbox}\\
    \begin{adjustbox}{max width=1.0\textwidth,center}
    \includegraphics[scale=0.35]{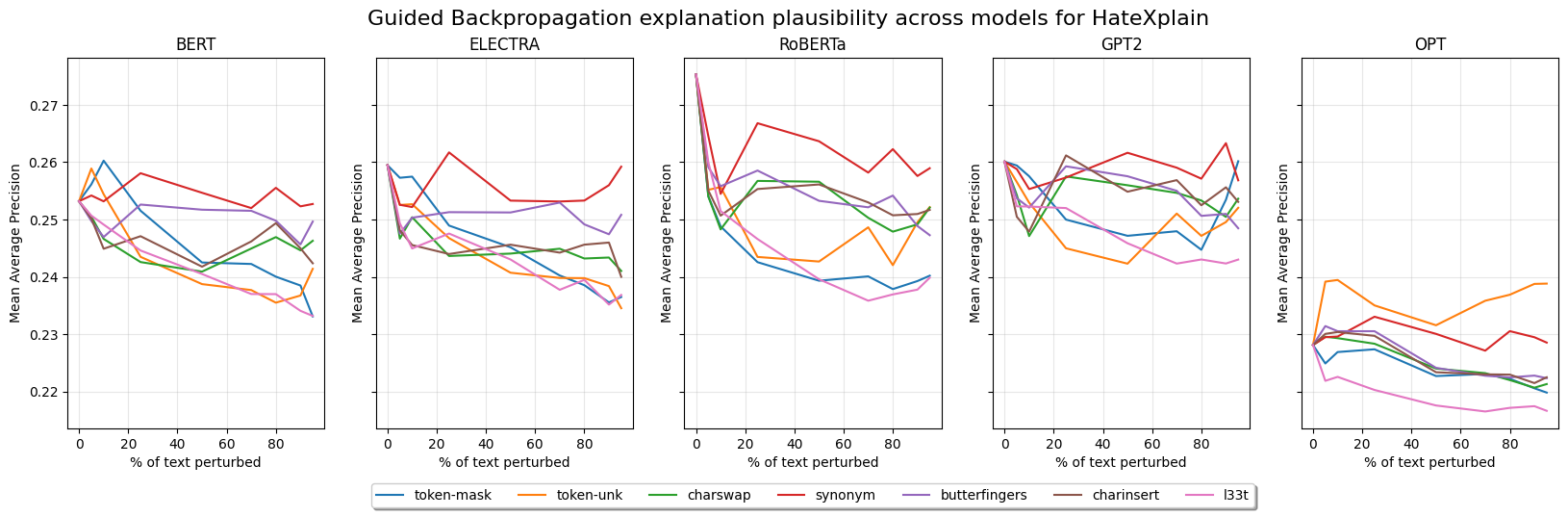}\end{adjustbox}\\
    \begin{adjustbox}{max width=1.0\textwidth,center}
    \includegraphics[scale=0.35]{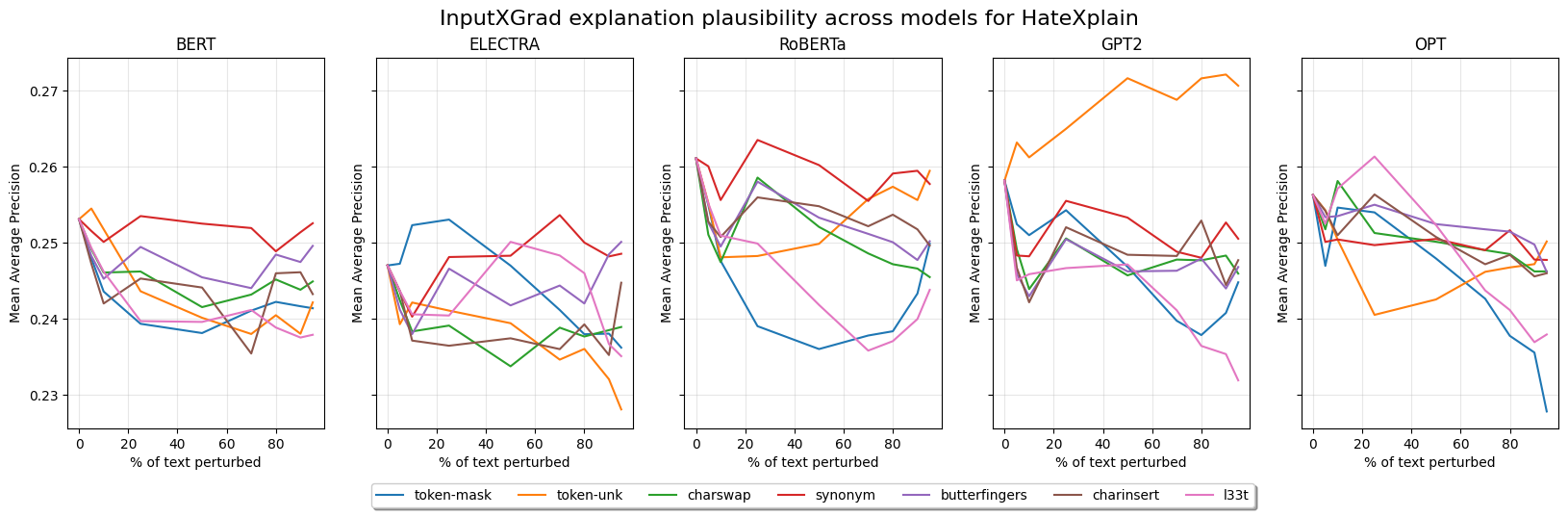}\end{adjustbox}\\
    
    \caption{We show the differential effect of increasing levels of text perturbation on the explanation plausiblity of all saliency maps \textbf{HateXplain} dataset. Values are averaged over all hierarchies.}
    \label{fig:TASK_DIFF_HX_COH}
\end{figure*}

We show the results of our investigations into the differential effect of perturbation types (see \S\ref{sec:RQ1_res}) across our datasets and models. We present the effect of perturbations across models on accuracy and predictive and epistemic uncertainty for the SST-2 dataset in Figure \ref{fig:TASK_DIFF_SST}, and the effect of perturbations on explanation plausibility in Figure \ref{fig:TASK_DIFF_SST_COH}. Similarly, we provide model-specific graphs for the SemEval dataset in Figures \ref{fig:TASK_DIFF_SEM} and \ref{fig:TASK_DIFF_SEM_COH}, and the HateXplain dataset in Figures \ref{fig:TASK_DIFF_HX} and \ref{fig:TASK_DIFF_HX_COH}. \textbf{Results:} We typically see a steep decrease in accuracy with increasing perturbation across all perturbations, models, datasets. Typically, we also see an increase in predictive and epistemic uncertainty; though, we see some exceptions with the HateXplain datasets, and this increase does not always correspond to the performance decrease. We generally see a decreasing trend in explanation plausibility with increasing perturbation, but this relationship is not as strong as the other observed metrics. Typically, this decrease is steepest with Integrated Gradients and all BERT explanations. Interestingly, we do not see SmoothGrad explanation plausibility change with perturbation with OPT and GPT2, which is related to the robustness of the combination as we see in Appendix \ref{app:RQ3_all} and the poor initial plausibility score.\\ 

Across all datasets, we find similar behaviour between special token replacements (\verb|token-unk| and \verb|token-mask|) as well as between character-level changes (\verb|charswap|, \verb|charinsert|, \verb|butterfingers|). \verb|synonym| and \verb|butterfingers| typically have the smallest effect on all measures. Interestingly, \verb|l33t| has a very task-dependent effect: For SemEval, it has a moderate effect on all model outputs. In HateXplain, it has a very strong negative effect on model performance and explanation plausibility, yet decreases model uncertainty. Generally, we see increasing uncertainty with increasing levels of perturbation for all models and noise types as well as decreasing accuracy. Typically, perturbations decrease accuracy at a similar rate as they increase uncertainty, except in the case of \verb|l33t|,  \verb|UNK|, and \verb|MASK|. With SST-2, \verb|l33t|,  \verb|UNK|, and \verb|MASK| perturbations most impact all models' accuracy, yet we do not see this reflected in the uncertainty curves. Similarly, for HateXplain, these perturbations reduce uncertainty for BERT, ELECTRA, RoBERTa, and OPT. Overall, GPT2 outputs much greater predictive and epistemic uncertainty relative to the other base models, and RoBERTa shows only slight increase in uncertainty with increased perturbation, even when accuracy is just as perturbed as other models (SST-2), suggesting a tendency for over-confidence.

\subsection{Robustness across perturbation types}
\label{app:RQ3_all}
We look at model-level differences in saliency map robustness across datasets at low levels of perturbation in Figure \ref{fig:RQ3-og_all}. We look at robustness at high levels of perturbation in Appendix \ref{app:50_corr}. \textbf{Results}: Typically, we see the greatest overall robustness across all perturbation types for Integrated Gradient. However, GPT2 and OPT typically has the greatest robustness with SmoothGrad. Guided backpropogation is typically very unrobust for all models, save for OPT, where it shows surprising robustness. For all models, we see a similar shape on the radar plot appear by Guided backpropagation, Integrated Gradients and InputXGrad, which is consistent for each model across datasets. Saliency maps on BERT typically show decreased robustness to \verb|l33t| perturbation, and increased robustness to \verb|butterfingers| and \verb|synonym|. For RoBERTa and GPT2, we see a decreased robustness to \verb|UNK| and \verb|l33t| perturbations, whereas robustness is lowest to \verb|MASK| perturbations for ELECTRA.  OPT shows consistent poor robustness to \verb|l33t|, and relatively low robustness to \verb|UNK| and \verb|MASK|, though the extent of this changes between tasks.

\begin{figure*}
    \centering
    \begin{adjustbox}{max width=1.0\textwidth,center}
    \includegraphics[scale=0.35]{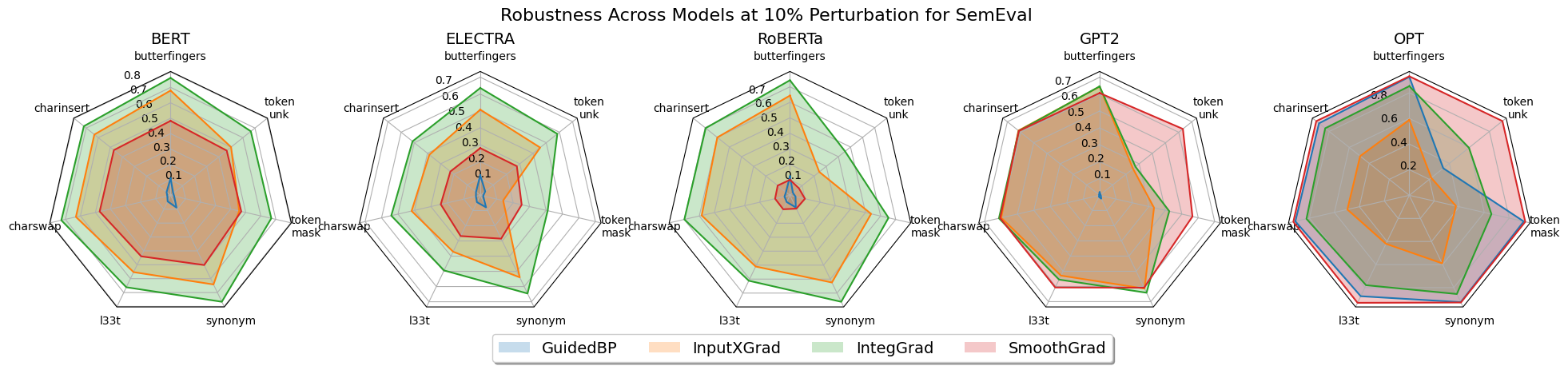}\end{adjustbox}\\
    \begin{adjustbox}{max width=1.0\textwidth,center}
    \includegraphics[scale=0.35]{RQ3-og_sst.png}\end{adjustbox}\\
    \begin{adjustbox}{max width=1.0\textwidth,center}
    \includegraphics[scale=0.35]{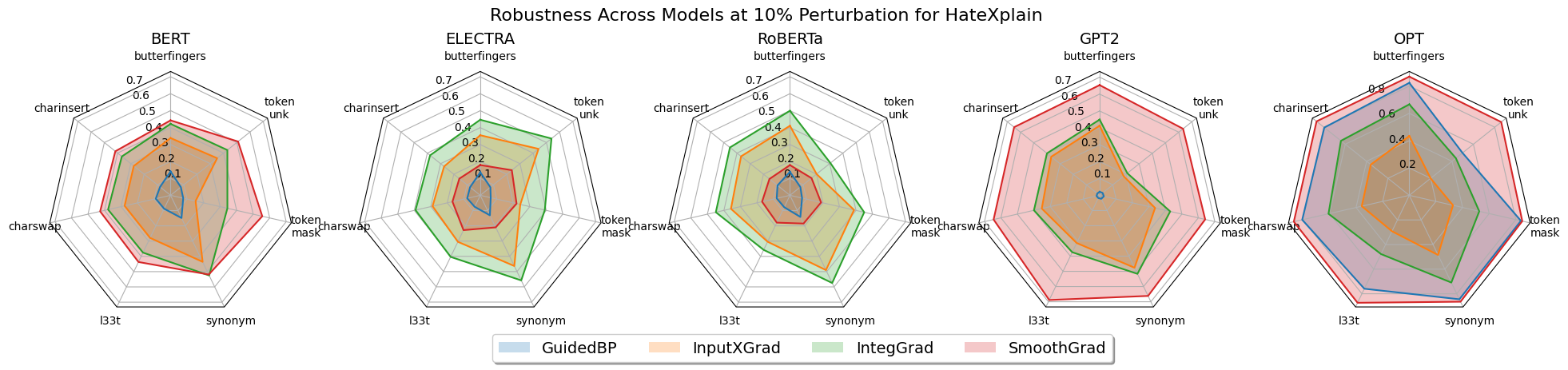}\end{adjustbox}\\
    \caption{We show the robustness across models of our saliency maps at low levels of perturbation across different tasks}
    \label{fig:RQ3-og_all}
\end{figure*}

\section{Extra investigations}
\subsection{Human-Random vs Human-Strategic}
\label{app:RvsS}
\begin{figure*}
    \centering
    \begin{adjustbox}{max width=1.0\textwidth,center}
    \includegraphics[scale=0.38]{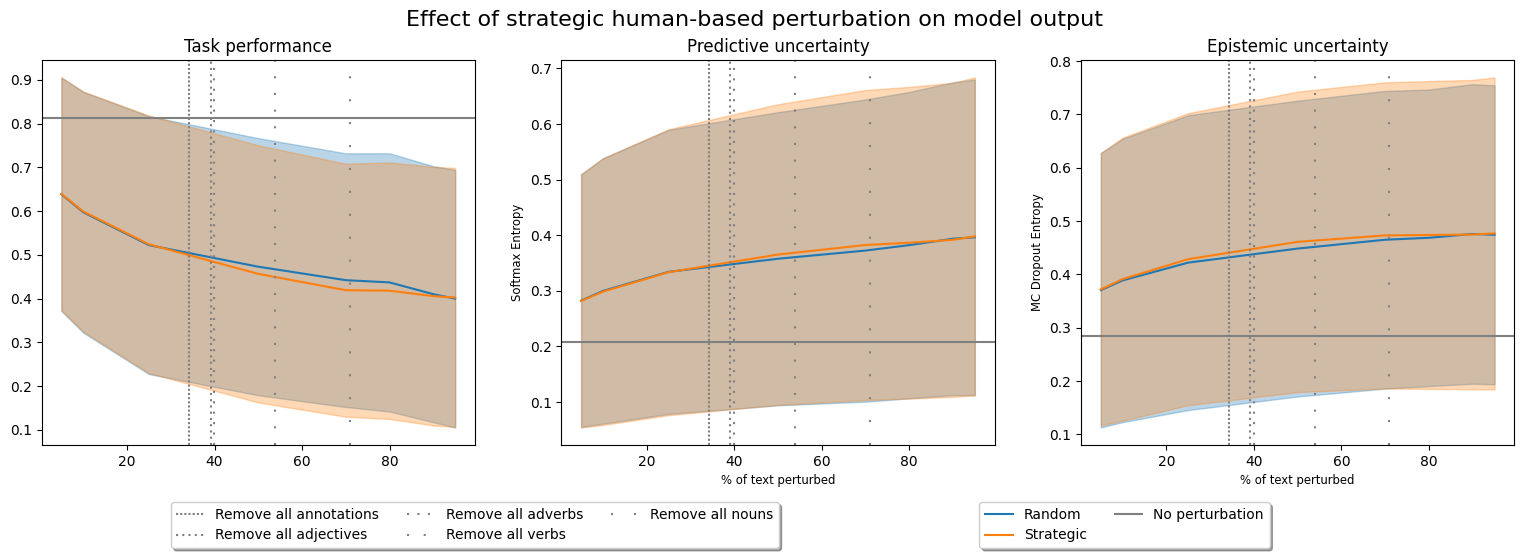}%
    \end{adjustbox}
    \caption{We compare the effect of two different methods of human-based perturbation on model accuracy, confidence and explanation plausibility. Human-Random randomly perturbs tokens after all annotated tokens are perturbed. Human-Strategic preferentially perturbs tokens based on their POS. Vertical lines denote the average location of strategy shift for the Human-Strategic perturbation hierarchy.}
    \label{fig:RQ1rvs}
\end{figure*}

To assess the efficacy of our human-strategic approach (and if POS tag-level perturbations affect model performance), we compare human-random and human-strategic perturbation in Figure \ref{fig:RQ1rvs}, and denote the average location of a change in strategy with a dotted line. \textbf{Results}: We can see that POS-hierarchied perturbation does adversely affect model performance and uncertainty. However, we find that after all adjectives, adverbs, verbs, and nouns have been perturbed, further perturbation does not show any increasing impact on model performance or uncertainty until the text is nearly completely perturbed. 

\subsection{Saliency map correlation to noise}
\label{app:noise_corr}
\begin{figure*}
    \centering
    \begin{adjustbox}{max width=1.0\textwidth,center}
    \includegraphics[scale=0.35]{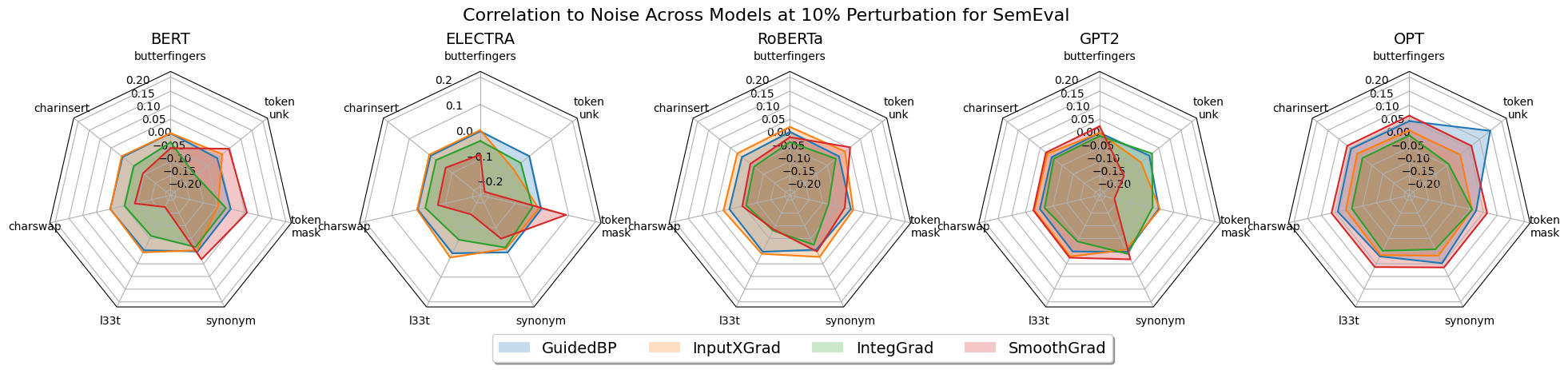}\end{adjustbox}\\
    \begin{adjustbox}{max width=1.0\textwidth,center}
    \includegraphics[scale=0.35]{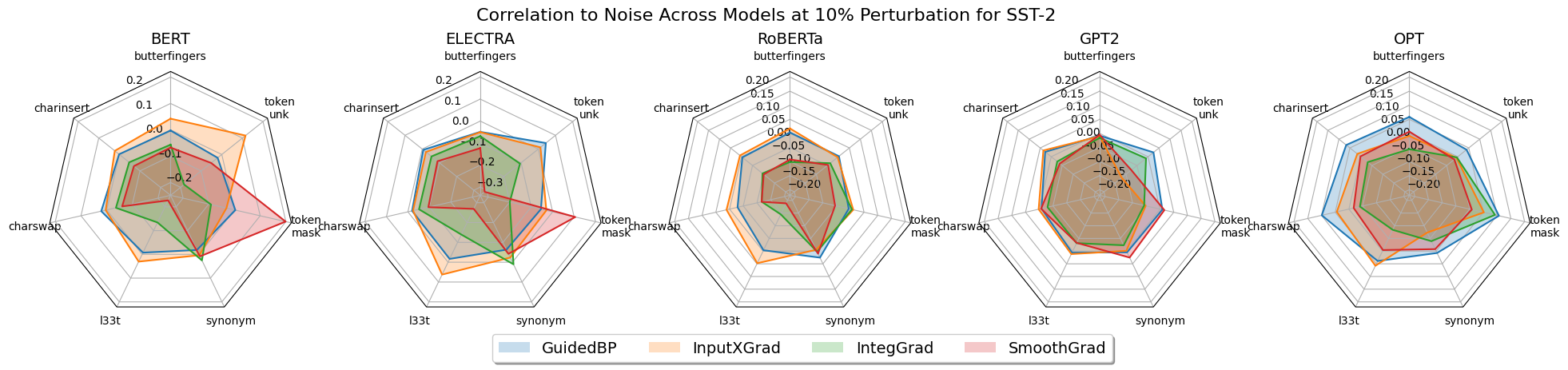}\end{adjustbox}\\
    \begin{adjustbox}{max width=1.0\textwidth,center}
    \includegraphics[scale=0.35]{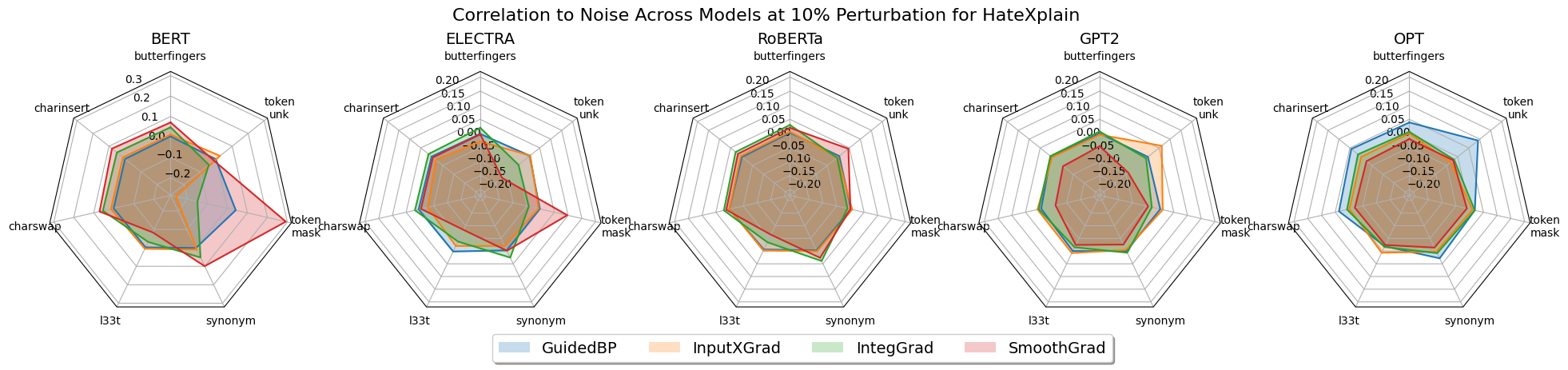}\end{adjustbox}\\
    \caption{We show the correlation to noise across models of our saliency maps at low levels of perturbation across different tasks}
    \label{fig:RQ3-noise_all}
\end{figure*}

To assess if decreased robustness of a saliency map technique to a particular perturbation type stems from attribution of saliency to the perturbed input, we assess the Pearson correlation of the output saliency map to the perturbed tokens, and visualize the output across dataset and model in a radar plot in Figure \ref{fig:RQ3-noise_all}.
\textbf{Results}: While we see equivalent lack of correlation to all types of noise for InputXGrad and GuidedBP saliency maps, SmoothGrad shows differing behaviour according to model type. For most models, SmoothGrad shows a slight negative correlation to \verb|l33t| and \verb|UNK| tokens; however, SmoothGrad does not show this particular aversion to \verb|UNK| with RoBERTa and it does not show a particular aversion to \verb|l33t| with GPT2. Furthermore, SmoothGrad applied on ELECTRA shows a consistent aversion to \verb|UNK| and \verb|l33t| tokens. All correlation values are very low, with a magnitude under 0.3. This behaviour for SmoothGrad may stem from its regularization process, and may also give some indication of model instability or stability, as these specific perturbations also have a strong detrimental effect on model performance. Ultimately, no model and saliency map combinations appear to preferentially attribute saliency to any type of perturbed tokens/words. %

\subsection{Saliency map robustness at high levels of noise}
\label{app:50_corr}
To assess saliency map robustness at high levels of noise, we present the same investigation performed in \S\ref{sec:RQ3_res} but with $\alpha=.5$ in Figure \ref{fig:RQ3-50_og}. 
\textbf{Results}: We see similar patterns as those described in Appendix \ref{app:RQ3_all}, but overall lower robustness. One exception is SmoothGrad on the larger language models (GPT2 and OPT) and Guided Backpropagation for OPT, which still seem to show high general robustness. However, though the general patterns of the saliency map\textsubscript{$\alpha=0.0$} is preserved, we can see in Figures \ref{fig:TASK_DIFF_SST_COH}, \ref{fig:TASK_DIFF_SEM_COH}, \ref{fig:TASK_DIFF_HX_COH}, the quality of the original saliency maps are typically quite low, in terms of agreement to human annotations. Similarly to \S\ref{sec:RQ3_res}, we can see that robustness is typically lower to \verb|l33t| and \verb|UNK| perturbations for BERT and GPT2, \verb|l33t| and \verb|MASK| for ELECTRA, and \verb|l33t|, \verb|MASK| and \verb|UNK| for OPT. %

 \begin{figure*}
    \centering
    \begin{adjustbox}{max width=1.0\textwidth,center}
    \includegraphics[scale=0.35]{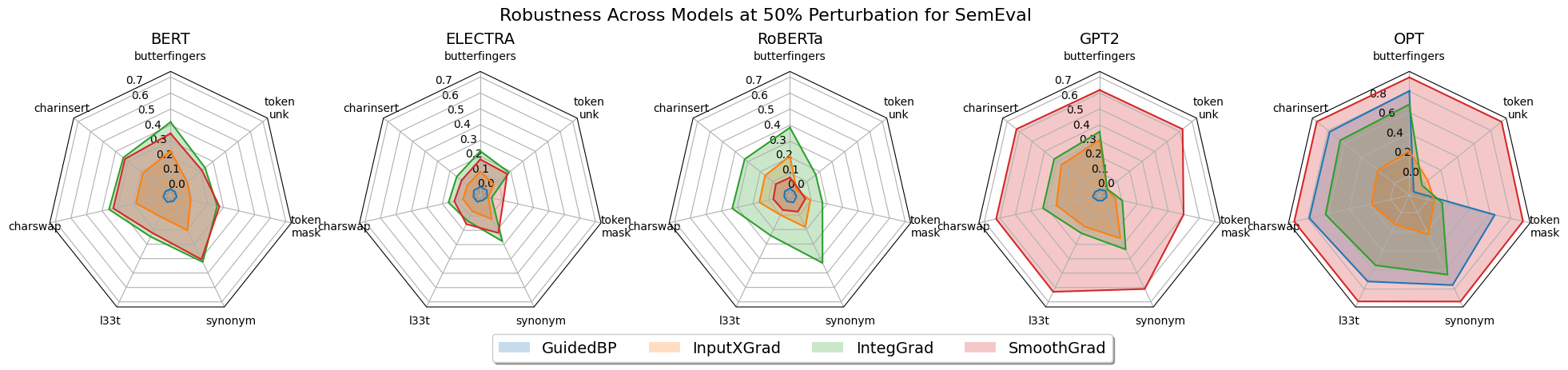}\end{adjustbox}\\
    \begin{adjustbox}{max width=1.0\textwidth,center}
    \includegraphics[scale=0.35]{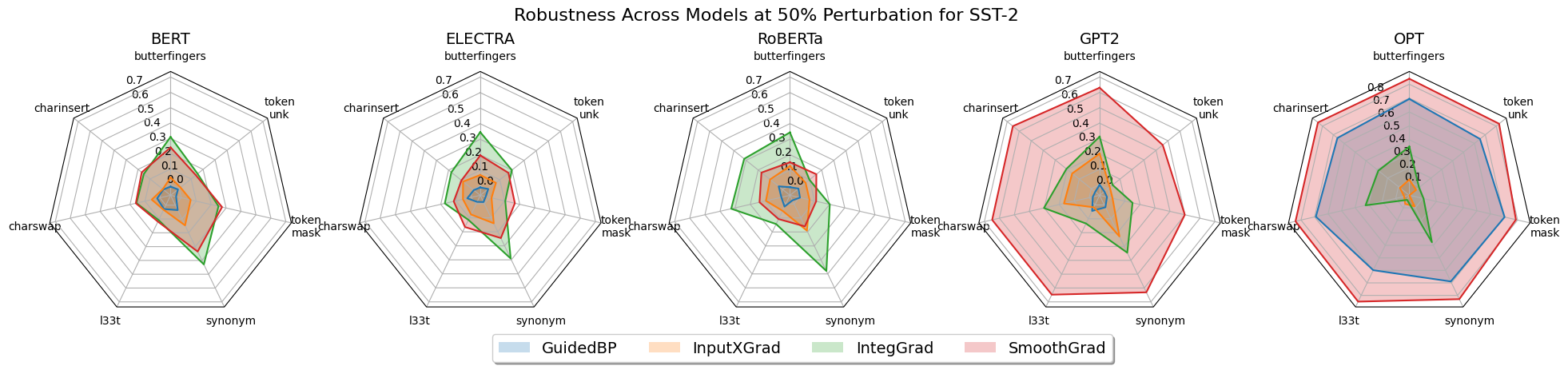}\end{adjustbox}\\
    \begin{adjustbox}{max width=1.0\textwidth,center}
    \includegraphics[scale=0.35]{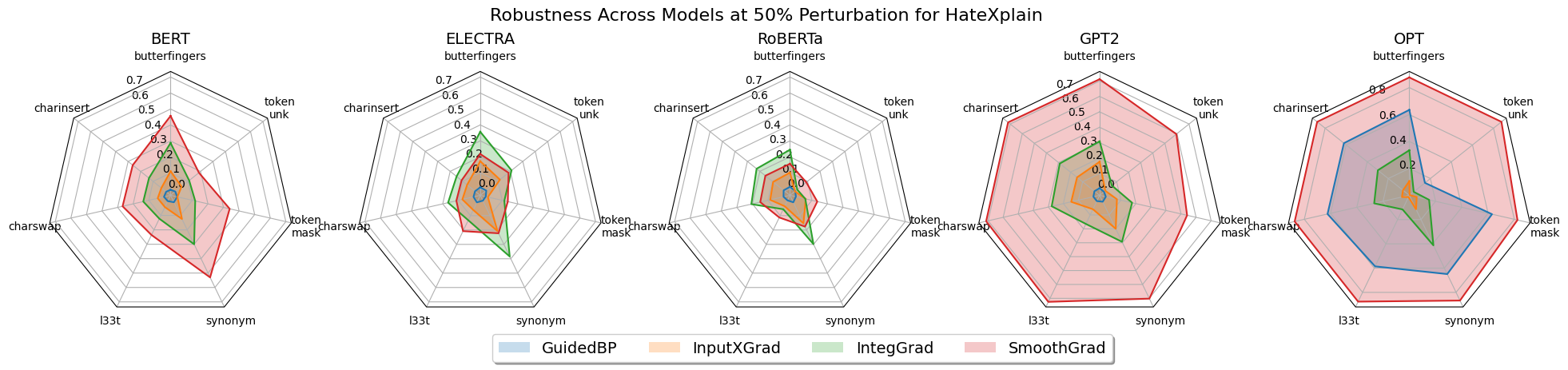}\end{adjustbox}\\
    \caption{We show the robustness of our saliency maps at high levels of perturbation ($\alpha=0.50$) across different tasks}
    \label{fig:RQ3-50_og}
\end{figure*}

\subsection{Uncertainty and explanation plausibility at high levels of perturbation}
\label{sec:unc_high}
We investigate the correlation between explanation plausibility and our two uncertainty measures at very high levels of perturbation ($\alpha\in\{.90,.95\}$) in Table \ref{tab:unc_high}, to assess if the previously observed relationship breaks down after salient tokens are removed. In this comparison, we also include incorrectly guessed datapoints. \textbf{Results:} In SST-2, which has no noise in its training data, we continue to observe a moderately negative relationship between uncertainty and explanation plausibility. SemEval, which is an easier task than HateXplain, seems to conserve a very weak positive relationship between uncertainty and explanation plausibility across models and attribution methods. However, for HateXplain, this correlation disappears (ca. 0.0), which suggests that the model can no longer identify salient tokens.

\begin{table*}[]
    \begin{adjustbox}{max width=1.0\textwidth,center}
\begin{tblr}{
  row{even} = {r},
  row{3} = {r},
  row{5} = {r},
  row{7} = {r},
  row{9} = {r},
  row{11} = {r},
  row{13} = {r},
  row{15} = {r},
  row{17} = {r},
  cell{1}{3} = {c=4}{c},
  cell{1}{7} = {c=4}{c},
  cell{3}{1} = {r=5}{},
  cell{8}{1} = {r=5}{},
  cell{13}{1} = {r=5}{},
  vline{4,8} = {1}{},
  vline{3,7,11} = {2-17}{},
  hline{1,18} = {-}{0.08em},
  hline{2-3,8,13} = {-}{0.05em},
}
                    &                        & \textbf{Predictive Uncertainty} &                         &                         &                         & \textbf{Epistemic Uncertainty} &                         &                         &                 \\
\textbf{Dataset}    & \textbf{Model}         & \textbf{GBP}                    & \textbf{IXG}            & \textbf{IG}             & \textbf{SG}             & \textbf{GBP}                   & \textbf{IXG}            & \textbf{IG}             & \textbf{SG}     \\
\textbf{SST-2}      & \textbf{BERT}          & -0.016                          & 0.020                   & -0.015                  & \textbf{0.092}          & \textbf{-0.162}                & -0.100                  & -0.089                  & -0.011          \\
                    & \textbf{ELECTRA}       & \textbf{-0.122}                 & -0.114                  & -0.048                  & -0.032                  & \textbf{-0.308}                & -0.289                  & -0.160                  & -0.151          \\
                    & \textbf{RoBERTa}       & \textbf{-0.169}                 & -0.123                  & -0.153                  & -0.130                  & \textbf{-0.315}                & -0.254                  & -0.244                  & -0.178          \\
                    & \textbf{\textbf{GPT2}} & \textbf{\textbf{-0.075}}        & -0.017                  & -0.070                  & -0.016                  & -\textbf{0.159}                & -0.100                  & -0.096                  & -0.048          \\
                    & \textbf{OPT}           & -0.013                          & -0.013                  & \textbf{-0.054}         & \textbf{-0.054}         & \textbf{-0.070}                & -\textbf{0.070}         & -0.020                  & -0.020          \\
\textbf{SemEval}    & \textbf{BERT}          & 0.088                           & \textbf{0.103}          & 0.088                   & \textbf{0.103}          & 0.089                          & \textbf{0.104}          & 0.087                   & 0.103           \\
                    & \textbf{ELECTRA}       & \textbf{0.103}                  & 0.096                   & \textbf{0.103}          & 0.096                   & \textbf{0.105}                 & 0.097                   & 0.104                   & 0.097           \\
                    & \textbf{RoBERTa}       & \textbf{0.106}                  & \textbf{0.106}          & \textbf{0.106}          & \textbf{0.106}          & \textbf{0.108}                 & 0.106                   & 0.104                   & 0.104           \\
                    & \textbf{\textbf{GPT2}} & 0.064                           & \textbf{\textbf{0.083}} & 0.065                   & \textbf{\textbf{0.083}} & 0.065                          & \textbf{\textbf{0.085}} & 0.065                   & 0.084           \\
                    & \textbf{\textbf{OPT}}  & \textbf{0.074}                  & \textbf{0.074}          & 0.073                   & 0.073                   & 0.067                          & 0.067                   & \textbf{0.076}          & \textbf{0.076}  \\
\textbf{HateXplain} & \textbf{BERT}          & -0.049                          & \textbf{-0.078}         & -0.049                  & \textbf{-0.078}         & -0.040                         & -0.060                  & -0.041                  & \textbf{-0.064} \\
                    & \textbf{ELECTRA}       & -0.054                          & -0.084                  & -0.061                  & \textbf{-0.091}         & -0.033                         & -0.059                  & \textbf{-0.060}         & -0.090          \\
                    & \textbf{RoBERTa}       & -0.021                          & -0.054                  & -0.023                  & \textbf{-0.055}         & -0.009                         & -0.036                  & -0.020                  & \textbf{-0.052} \\
                    & \textbf{\textbf{GPT2}} & 0.134                           & 0.090                   & \textbf{\textbf{0.140}} & 0.094                   & 0.126                          & 0.097                   & \textbf{\textbf{0.134}} & 0.092           \\
                    & \textbf{\textbf{OPT}}  & \textbf{0.019}                  & \textbf{0.019}          & 0.003                   & 0.003                   & -\textbf{0.025}                & \textbf{-0.025}         & 0.005                   & 0.005           
\end{tblr}
\end{adjustbox}
\caption{The Spearman Rank Correlation between explanation plausibility (MAP) and both measures of uncertainty across model, dataset and saliency map at high levels of perturbation ($\alpha\in\{0.90,0.95\}$) All datapoints (correctly and incorrected guessed) are included. We bold the saliency map with the strongest correlation for each comparison.}
\label{tab:unc_high}
\end{table*}

\end{document}